\documentclass{naturep}

\usepackage{caption}
\usepackage{amssymb}
\usepackage{amsmath}
\usepackage{graphicx}
\usepackage{algorithm}
\usepackage{algpseudocode}
\usepackage{ltablex}
\usepackage{bbm}
\usepackage{lineno}
\usepackage[margin=0.6in]{geometry}
\usepackage{ragged2e}
\usepackage{setspace}
\usepackage{longtable}
\usepackage[hidelinks]{hyperref}
\usepackage{anyfontsize}
\usepackage{float}
\usepackage{booktabs}
\usepackage{multirow}
\usepackage[dvipsnames]{xcolor}
\usepackage{blindtext}
\usepackage{caption}
\usepackage{chngcntr}
\usepackage{subcaption}
\usepackage{cleveref}
\usepackage{arydshln}
\usepackage{enumitem, kantlipsum}
\usepackage{caption}
\usepackage{afterpage}
\usepackage{tcolorbox}
\tcbuselibrary{skins, breakable}
\definecolor{promptbg}{RGB}{245,247,250}
\definecolor{promptborder}{RGB}{100,130,180}

\newcommand{\ppg}[1]{\,\textcolor{PineGreen}{\scriptsize$_{#1}$}}
\newcommand{\ppr}[1]{\,\textcolor{BrickRed}{\scriptsize$_{#1}$}}
\newcommand{\ppz}[1]{\,\textcolor{gray}{\scriptsize$_{#1}$}}

\makeatletter
\let\saved@includegraphics\includegraphics
\AtBeginDocument{\let\includegraphics\saved@includegraphics}

\makeatother

\title{\begin{flushleft}\begin{spacing}{1}Unified Multi-Foundation-Model Slide Representation for Pan-Cancer Recognition and Text-Guided Tumor Localization\end{spacing}\end{flushleft}}
\author{}

\begin{document}

\maketitle

\begin{spacing}{1}

\vspace{-16mm}
\noindent
Tianyang Wang$^{1,2,*}$, Ziyu Su$^{1}$, Abdul Rehman Akbar$^{1,2}$, Usama Sajjad$^{1,2}$, 
Lina Gokhale$^{1}$, Charles Rabolli$^{1}$, Wei Chen$^{1}$, Anil Parwani$^{1}$, and Muhammad Khalid Khan Niazi$^{1,2}$
\end{spacing}

\vspace{-9mm}
\begin{spacing}{1.4}
\begin{affiliations}

 \item Department of Pathology, College of Medicine, The Ohio State University Wexner Medical Center, Columbus, OH, USA

 \item Department of Biomedical Engineering, The Ohio State University, Columbus, OH, USA

 \par\noindent \textbf{Corresponding author}: Tianyang Wang (Tianyang.Wang@osumc.edu)

\end{affiliations}
\end{spacing}

\vspace{-8mm}
\begin{spacing}{1}

\noindent \textbf{Abstract}

The expanding ecosystem of pathology foundation models has produced powerful but fragmented tile-level representations, limiting their use in clinical tasks that require unified slide-level reasoning and interpretable linkage to clinically meaningful information. We present ASTRA, a pan-cancer framework that integrates heterogeneous foundation-model representations into a shared slide-level representation space and semantically grounds that space using structured pathology annotation fields, including classification category, cancer type, and anatomic site. ASTRA combines sparse mixture-of-experts contextualization, masked multi-model reconstruction, and contrastive alignment to structured pathology prompts to learn slide representations that support 4-category classification, 3-class solid tumor typing, 16-class cancer typing, and text-guided tumor localization without pixel-level supervision. Developed on a CHTN cohort of 10,359 whole-slide images (WSIs) spanning 16 tumor types, ASTRA consistently improves pan-cancer classification across four pathology foundation-model backbones, achieving up to 97.8\% macro-AUC for 4-category classification, 99.7\% for 3-class solid tumor typing, and 99.2\% for 16-class cancer typing. For tumor localization, ASTRA achieves a mean Dice of 0.897 on an annotated in-domain CHTN subset ($n=380$) spanning 16 cancer types and 0.738 on an external TCGA cohort ($n=1{,}686$) spanning four cancer types. These results demonstrate that minimal structured pathology annotation fields derived from slide-level metadata can provide effective semantic supervision for unified slide representation learning, enabling both pan-cancer prediction and weakly supervised tumor localization within a single framework.

\end{spacing}

\vspace{-4mm}
\begin{spacing}{1}
\newpage
\noindent\textbf{\large{Introduction}}

The digitization of pathology has transformed histopathologic assessment into a scalable computational paradigm, enabling quantitative analysis of whole-slide images (WSIs) in both research and clinical practice~\cite{baxi2022digital,niazi2019digital,rajpurkar2022ai,akbar2025learning}. Recent pathology foundation models further show that local tissue morphology can be encoded into rich and transferable tile-level representations without task-specific supervision~\cite{su2025streamline,chen2024uni,zimmermann2024virchow2,xu2024gigapath,lu2024conch,chen2026ranger}. These advances have substantially expanded what can be learned from individual image tiles. Yet many questions of clinical and biological interest are not defined at the level of an isolated tile, but at the level of the WSI, where diagnostically relevant signals are distributed across spatially organized tissue regions. Translating strong tile-level representations into slide-level representations that preserve local spatial context, remain interpretable, and support robust downstream inference across tasks and cohorts therefore remains a central challenge in computational pathology.

Tumor localization provides a stringent test of this broader problem. A single WSI often contains tumor, benign epithelium, stroma, necrosis, and background in complex spatial mixtures, and many downstream analyses depend on identifying which regions truly harbor tumor~\cite{choi20212020}. Fully supervised segmentation frameworks such as U-Net, nnU-Net, and HookNet can achieve strong performance when dense annotations are available~\cite{ronneberger2015u,isensee2021nnu,van2021hooknet}, but producing pathologist-delineated pixel-level tumor annotations at WSI scale is costly, labor-intensive, and subject to inter-observer variability~\cite{wang2022label,verghese2023computational}. These practical constraints have motivated weakly supervised approaches that infer tumor regions directly from WSI level labels~\cite{campanella2019clinical,lu2021data,ilse2018attention,lu2024conch,shao2021transmil}. However, when supervision is limited to coarse global labels alone, the resulting representations often lack the semantic specificity needed for precise localization and may generalize inconsistently across tumor types and cohorts.

This challenge is compounded by the growing ecosystem of pathology foundation models. Models such as UNI, Virchow2, GigaPath, and CONCH each produce strong tile-level representations~\cite{chen2024uni,zimmermann2024virchow2,xu2024gigapath,lu2024conch}. However, these representations are fundamentally diverse, as variations in training data, model architecture, and learning objectives lead each model to capture distinct aspects of tissue morphology and semantics. This heterogeneity creates an opportunity, because complementary information may be distributed across models, but it also complicates downstream analysis, where selecting a single backbone is often empirical and may leave useful signals unexploited. Recent studies have therefore started to combine embeddings from multiple pathology foundation models at either the tile or slide level, with evidence that unified representations can outperform individual encoders~\cite{ticon,runevic2025combining,chen2026histomet}. Still, most existing strategies treat multi-model integration largely as feature fusion, without explicitly modeling local spatial structure or connecting the resulting slide representation to clinically grounded supervision.

Semantic grounding offers a promising route to bridge this gap. Recent vision-language pathology models, including TITAN~\cite{ding2025multimodal} and CONCH~\cite{lu2024conch}, have shown that aligning slide representations with pathology text supports strong zero-shot retrieval and classification, suggesting that text alignment also enables semantically informed localization without dense annotations. However, these approaches often depend on curated multimodal corpora that extend beyond routine clinical workflows. For example, TITAN~\cite{ding2025multimodal} uses multi-stage visual--language supervision, combining ROI-level alignment based on large pathology regions with synthetic PathChat-generated captions and WSI-level alignment based on slide--report pairs \cite{lu2023foundational}. Similarly, CONCH is pretrained on a large histopathology image-caption corpus constructed from educational resources and PubMed Central Open Access articles using an automated curation pipeline. These studies demonstrate the potential of semantic alignment, but they also rely on specialized multimodal data construction or report-level text that is not routinely available across institutions. It therefore remains unclear whether structured pathology annotation fields available through routine pathology data curation workflows, such as classification category, cancer class, and anatomic site, are sufficient to ground slide representations for spatially resolved tumor inference. More broadly, a unified framework is still lacking that can reconcile heterogeneous pathology foundation-model embeddings, preserve local spatial context, and semantically ground the resulting representations using only such structured annotation fields.

To address these challenges, we developed ASTRA, a pan-cancer framework for slide representation learning with semantic grounding. ASTRA unifies representations from multiple pathology foundation models, models local tissue context with a spatially aware sparse mixture-of-experts encoder, and aligns slide representations with structured pathology prompts derived from routinely available annotation fields. In contrast to prior vision-language approaches that rely on paired free-text reports or curated image-caption corpora, ASTRA uses structured pathology fields, including classification category, cancer type, and anatomic site, as scalable semantic supervision. We evaluated the learned representation on slide-level classification tasks of varying granularity and additionally examined whether the same semantically grounded representation could support text-guided tumor localization without pixel-level supervision.

\begin{center}
\includegraphics[width=\textwidth,height=0.98\textheight,keepaspectratio]{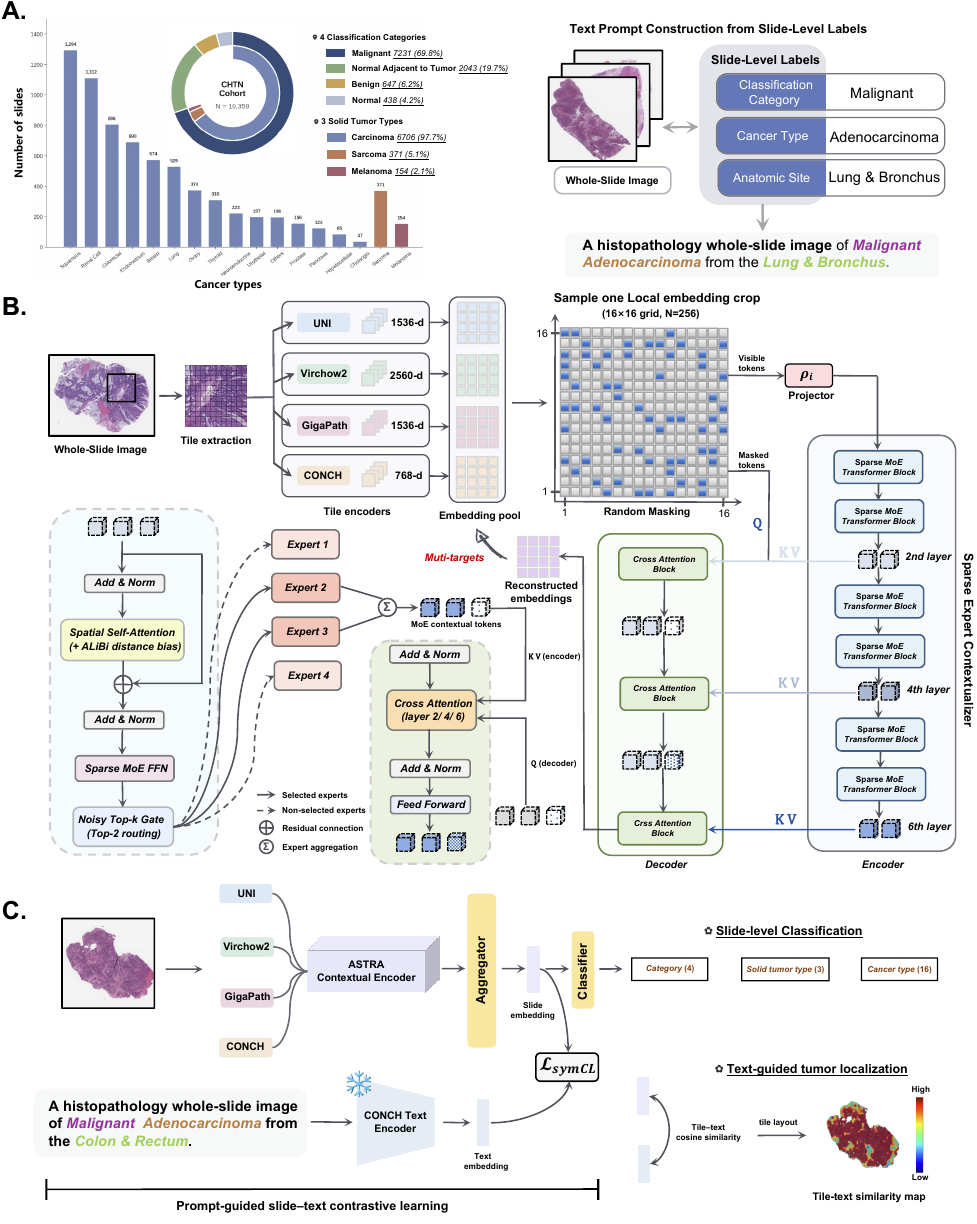}
\end{center}
\captionsetup[table]{position=bottom}
\captionsetup[figure]{labelfont=bf,textfont=normalfont}

\captionof{figure}{\textbf{Overview of ASTRA.}
\textbf{(A)} Composition of the CHTN pan-cancer cohort used to develop ASTRA, comprising 10,359  WSIs across 16 tumor types. Each slide is associated with structured pathology annotation fields  (classification category, cancer type, and anatomic site), which serve as semantic supervision for slide-level representation learning.
\textbf{(B)} ASTRA pretraining. WSIs are encoded by four pathology foundation models to form a shared embedding pool. Local spatial crops of tile embeddings are sampled and partially masked, and contextualized using a sparse mixture-of-experts (MoE) transformer. A hierarchical cross-attention decoder reconstructs embeddings from multiple foundation models, encouraging a unified representation that captures spatial context and is predictive across heterogeneous embedding spaces.
\textbf{(C)} Downstream tasks and semantic supervision in ASTRA. Contextualized tile embeddings are aggregated into slide-level representations for pan-cancer slide-level classification. During training, slide-level representations are aligned with structured pathology prompts via contrastive learning to inject semantic supervision. For text-guided tumor localization, tile embeddings are compared with text embeddings independently of the classification head to produce tile--text similarity maps. Further details are described in the \textbf{Methods}.}
\label{fig:overview}

\vspace{4mm}
\begin{spacing}{1}

\vspace{6mm}
\noindent\textbf{\large{Results}}\\

\noindent\textbf{Pan-Cancer Slide Classification}\\ 
We evaluated whether ASTRA improves slide-level classification performance across heterogeneous pathology foundation models. Using Gated ABMIL for slide aggregation and a linear classifier, we compared five contextualization strategies across four backbones on three held-out classification tasks of increasing granularity: 4-category classification ($n=2{,}072$), 3-class major-group prediction ($n=1{,}402$), and 16-class cancer typing on malignant slides ($n=1{,}446$). Quantitative results are summarized in \textbf{Figure~\ref{fig:astra_results}a--c}, with full metric breakdowns in \textbf{Supplementary Tables~\ref{stab:cls_category4}--\ref{stab:cls_cancer16}}.

Comprehensive evaluations across various tasks and backbone architectures indicate that ASTRA variants consistently outperform the corresponding Raw baselines, with the most pronounced improvements observed in macro-AUC and balanced accuracy. In the four-class setting, ASTRA increases balanced accuracy by up to $3.6$ (e.g., CONCH v1.5: $73.2 \rightarrow 76.8$) and macro-AUC by up to $0.7$ ($96.7 \rightarrow 97.4$), alongside steady AUC gains of $0.3$--$0.6$ across other backbones. Overall, ASTRA variants achieve or surpass the Raw baseline in macro-AUC across all 12 backbone-task configurations. Furthermore, the full ASTRA model demonstrates superior performance in 11 of these 12 cases, with the sole exception being the GigaPath backbone applied to the 16-class cancer-typing task.

ASTRA (ISO) also generally surpasses TICON (ISO) in balanced accuracy. Representative absolute increases in the four-class setting include $3.6$ for CONCH v1.5 and $1.2$ for Virchow2. 
These results demonstrate that ASTRA yields substantial representational benefits even under isolated feature extraction conditions, independent of slide-level spatial context. The efficacy of multi-foundation-model pretraining is most striking in highly fine-grained analytical tasks. Parallel improvements in balanced accuracy (up to $2.2$) and macro-AUC (up to $0.6$) are observed in the three-class major-group task. During the 16-class cancer-typing task, ASTRA enhances balanced accuracy by a maximum of $4.0$ (GigaPath: $81.3 \rightarrow 85.3$) and maintains an increase of approximately $1.2$ across multiple backbones, while macro-AUC exhibits consistent gains of $0.2$--$0.4$. These findings underscore the critical advantage of integrating complementary morphological features from multiple encoders to capture fine-grained histopathological distinctions.

\noindent\textbf{Mechanistic analysis of expert routing}\\
To visualize how ASTRA routes tissue tiles through the Sparse MoE encoder, we extracted tile-level routing assignments from the final MoE block of the trained model. For each tile, we identified the highest-probability expert from the routing distribution and mapped these assignments back to the original whole-slide coordinates to generate a smoothed slide-level expert partition map (\textbf{Figure~\ref{fig:expert_routing_vis}, left}). To summarize the morphology associated with each expert, we selected high-confidence tiles with clear expert separation and displayed the top-ranked tissue tiles within each expert as representative exemplars (\textbf{Figure~\ref{fig:expert_routing_vis}, right}).

A board-certified pathologist reviewed these expert-specific tile groups and found that the routed tilees corresponded to recurrent and distinct histologic patterns rather than random mixtures of tissue appearances. Specifically, Expert~1 predominantly captured poorly differentiated malignant epithelial cells characterized by solid tumor architecture, nuclear pleomorphism, and a high nuclear-to-cytoplasmic ratio. Expert~2 was enriched for gland-forming adenocarcinoma structures with luminal organization and columnar tumor cells. Expert~3 primarily represented benign or near-normal glandular epithelium with preserved polarity and relatively uniform nuclei. Expert~4 captured stromal and microenvironmental components, including fibroblastic stroma, adipocytic tissue, vascular structures, and inflammatory infiltrates.

These observations indicate that ASTRA's Sparse MoE encoder learns structured and morphologically coherent routing patterns, with different experts specializing in recurrent histologic phenotypes. Such spatial specialization suggests that ASTRA preserves fine-grained tissue organization during contextualization, motivating us to examine whether this property can support accurate tumor localization under guidance from slide-specific pathology prompts derived from routine annotations alone, without pixel-level supervision.

\begin{center}
\includegraphics[width=\textwidth,height=0.98\textheight,keepaspectratio]{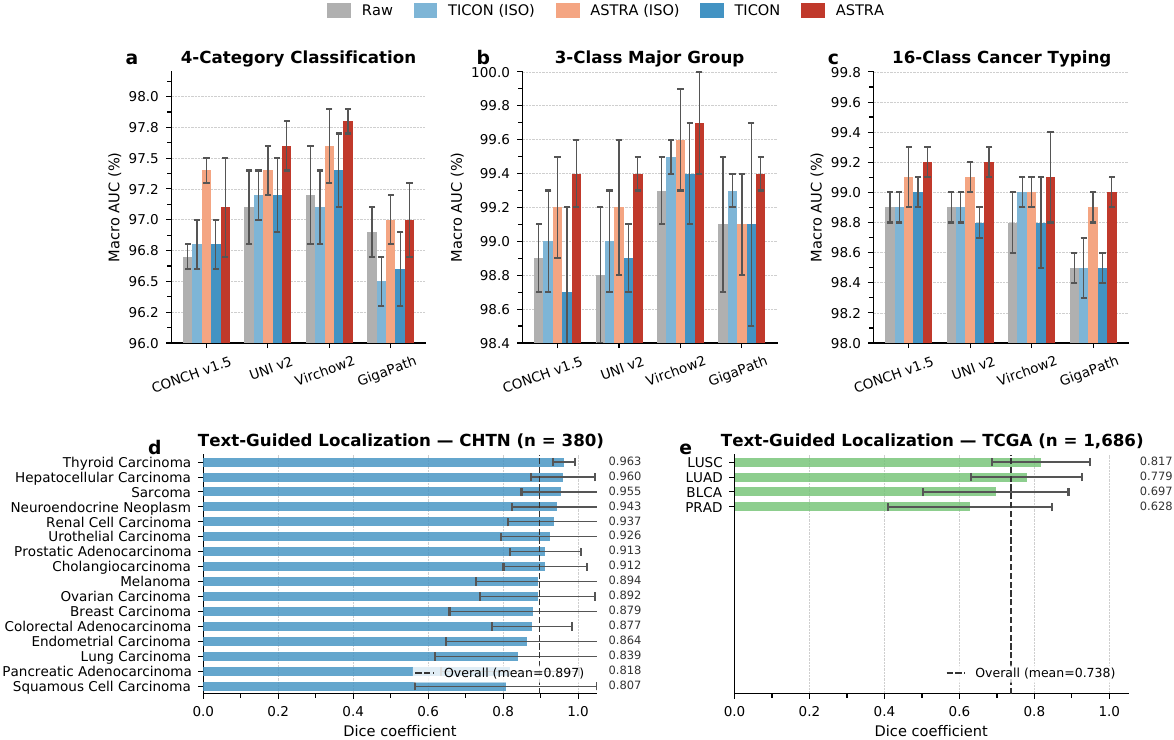}
\end{center}
\caption{\textbf{Pan-cancer performance of ASTRA in slide classification and text-guided tumor localization.}
\textbf{a--c}, Macro-averaged area under the receiver operating characteristic curve (AUC) across three slide-level classification tasks of increasing granularity: 4-class classification (\textbf{a}), 3-class major-group prediction (\textbf{b}), and 16-class cancer typing (\textbf{c}). Performance is compared across four pathology foundation-model backbones and five contextualization strategies. Error bars represent standard deviation over five independent random seeds. ASTRA variants generally outperform Raw and TICON baselines, with the strongest overall performance achieved by the full ASTRA model and peak results obtained with the Virchow2 backbone.
\textbf{d--e}, Text-guided tumor localization performance measured by Dice similarity coefficient on the in-domain annotated CHTN subset (\textbf{d}, $n=380$) and the external TCGA cohort (\textbf{e}, $n=1{,}686$), stratified by cancer type, including lung squamous cell carcinoma (LUSC), lung adenocarcinoma (LUAD), bladder urothelial carcinoma (BLCA), and prostate adenocarcinoma (PRAD). Error bars indicate standard deviation across slides. The dashed vertical line marks the overall mean Dice across all slides in the respective cohort.}
\label{fig:astra_results}

\begin{center}
\includegraphics[width=\textwidth]{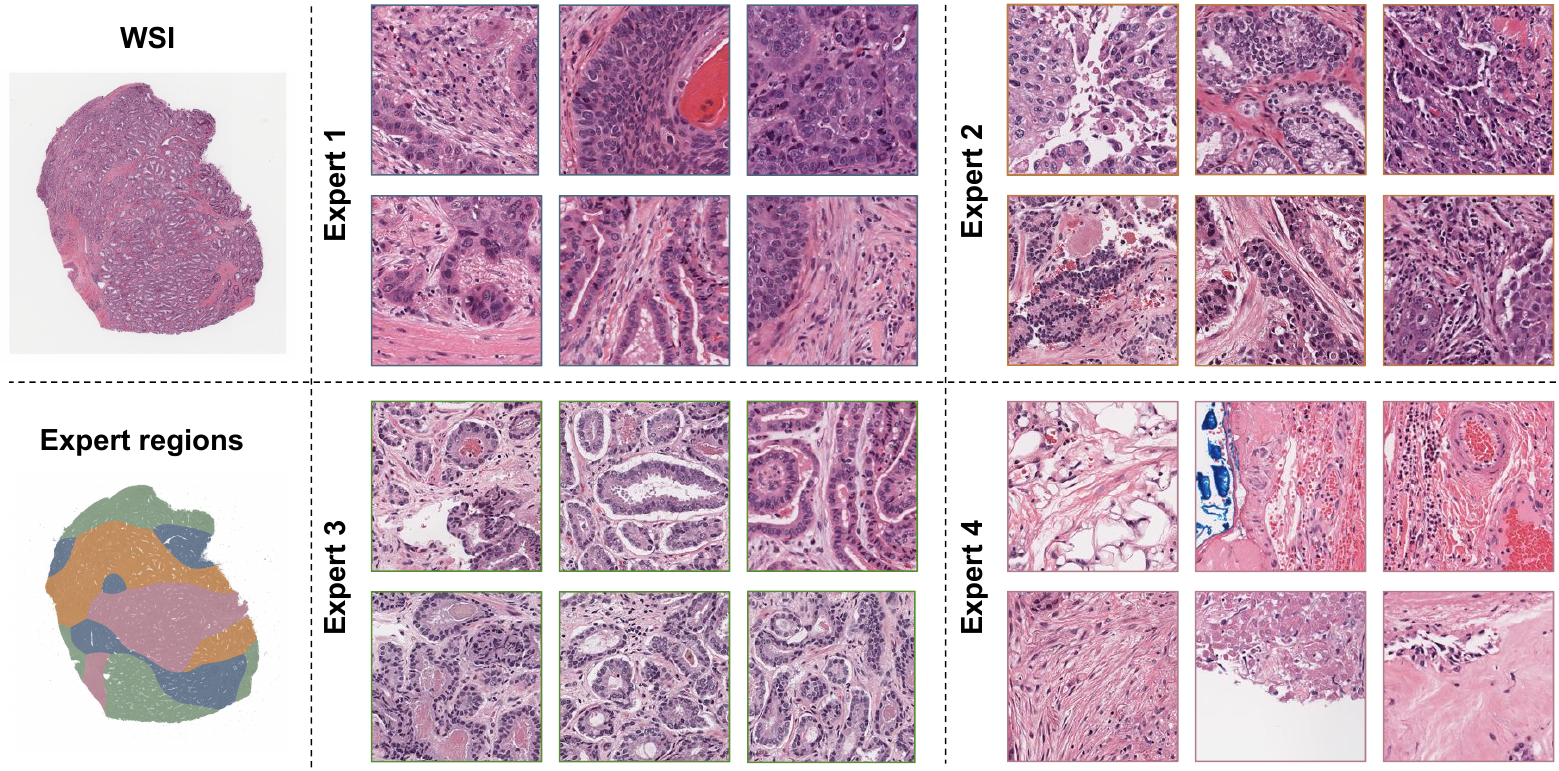}
\end{center}
\captionof{figure}{\textbf{Visualization of ASTRA expert routing and histologic specialization.}
Left, a representative WSI and the corresponding smoothed expert-region partition map derived from top-1 tile-level routing assignments in the final MoE block. Right, representative high-confidence tissue tiles assigned to each expert. Expert 1 predominantly captured poorly-differentiated carcinoma cells. Expert 2 was enriched for gland-forming carcinoma structures with moderate cytoarchitectural atypia. Expert 3 primarily represented well-formed tumor glands with low-grade cytoarchitectural atypia. Expert 4 captured tumor stromal and microenvironmental components.}
\label{fig:expert_routing_vis}

\noindent\textbf{Text-guided tumor localization}\\
We next evaluated whether ASTRA's semantically aligned representation could localize tumor regions without pixel-level supervision. Quantitative localization results are summarized in Figure~\ref{fig:astra_results}d--e, with full numerical breakdowns in \textbf{Supplementary Tables~\ref{stab:chtn_dice} and \ref{stab:tcga_dice}}. On the in-domain CHTN annotated subset ($n=380$, spanning all 16 cancer types), ASTRA achieved an overall mean Dice of $0.897$ and a median Dice of $0.969$, indicating high spatial overlap for most slides. Performance was strongest for thyroid carcinoma (mean Dice $0.963$), hepatocellular carcinoma ($0.960$), and sarcoma group ($0.955$), whereas lower but still competitive scores were observed for squamous cell carcinoma ($0.807$) and pancreas carcinoma ($0.818$).

Representative localization examples across all 16 CHTN cancer types are shown in \textbf{Figure~\ref{fig:chtn_localization}}. These examples show that the strong quantitative performance was accompanied by visually accurate recovery of tumor extent across diverse morphologic patterns, including compact nodular lesions such as hepatocellular and thyroid carcinoma, as well as more diffuse growth patterns such as squamous cell and urothelial carcinoma. Overall, the qualitative results are consistent with the broad Dice performance observed in the in-domain cohort.

We then evaluated generalization to an external TCGA cohort ($n=1{,}686$) in a zero-shot setting, without any fine-tuning on the target domain, using previously released slide-level tumor prediction maps as the external reference resource. ASTRA achieved an overall mean Dice of $0.738$, with the strongest performance on LUSC ($0.817$) and LUAD ($0.779$), intermediate performance on BLCA ($0.697$), and lower performance on PRAD ($0.628$). Although performance decreased relative to the in-domain CHTN subset, substantial localization accuracy was retained under zero-shot cross-cohort transfer, despite the absence of explicit pixel-level supervision during training. Representative qualitative examples from the external TCGA cohort are provided in~\textbf{Supplementary Figure \ref{sfig:tcga_luad}--\ref{sfig:tcga_blca}}.

\begin{center}
\includegraphics[width=\textwidth]{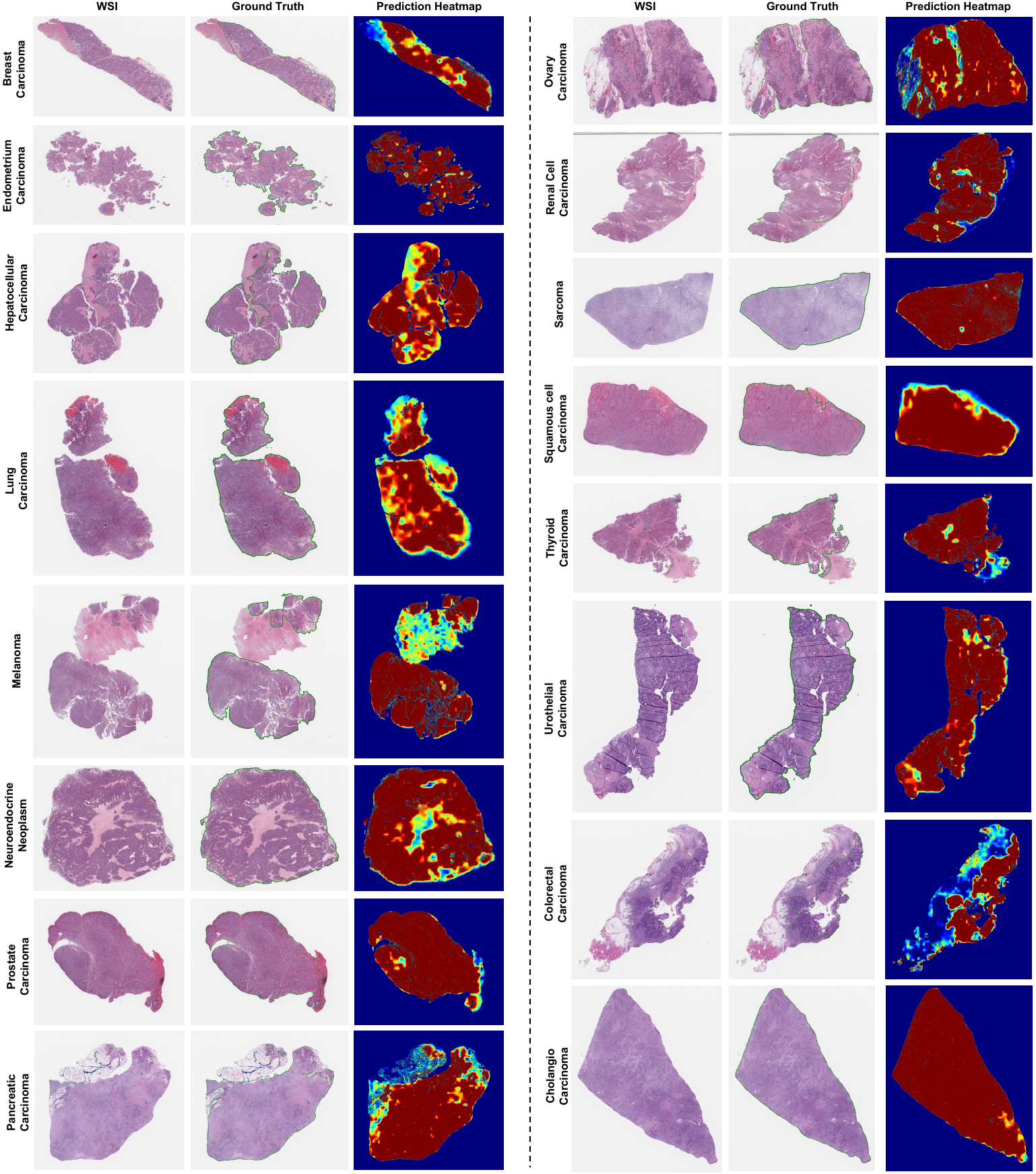}
\end{center}
\captionof{figure}{\textbf{Representative ASTRA tumor localization across 16 CHTN cancer types.}
Each row corresponds to one cancer type from the annotated CHTN subset. Columns show the H\&E whole-slide thumbnail (left), the ground-truth tumor contour overlaid in green (center), and the tile--text cosine similarity heatmap between ASTRA tile embeddings and the slide-specific pathology prompt (right). Warmer colors indicate higher similarity.}
\label{fig:chtn_localization}

\vspace{6mm}

\noindent\textbf{\large{Discussion}}\\

ASTRA addresses two related challenges in computational pathology: how to integrate heterogeneous pathology foundation-model embeddings at the slide level, and how to semantically ground those representations using structured pathology annotation fields. Across three classification tasks, ASTRA consistently outperformed raw foundation-model embeddings and prior contextualization baselines, with the strongest results observed with Virchow2. ASTRA also supported text-guided tumor localization without pixel-level supervision, achieving strong performance on the annotated CHTN subset and retaining substantial localization accuracy on external TCGA slides. These results show that multi-foundation-model contextualization and routine semantic alignment can support both slide-level prediction and spatial tumor inference.

The classification results highlight two points. First, the  largest gains were observed on the 3-class and 16-class tasks, where discrimination depends on subtle morphologic differences. This pattern suggests that different pathology foundation models capture complementary tissue embeddings. Second, ASTRA (ISO) frequently outperformed TICON (ISO), indicating that the benefit does not come only from multi-foundation-model pretraining. The sparse MoE contextualizer itself likely contributed to this improvement, consistent with the expert-routing analysis showing that different experts preferentially captured recurrent histologic phenotypes rather than arbitrary tissue partitions.

The localization results further clarify both the strengths and the limits of the learned representation. On the in-domain CHTN subset, strong Dice scores across 16 cancer types indicate that the aligned visual--text space preserves spatial information relevant to tumor extent. On TCGA, performance dropped but remained substantial without fine-tuning, indicating partial transfer under cohort shift. A major likely contributor to this reduction is that the reference labels were generated by another model rather than by pathologists, which may introduce label noise into the evaluation. Additional factors may include differences in staining, image acquisition, and cohort composition, as well as limited representation of some morphologies in the training set. More broadly, these findings show that structured pathology annotation fields, including classification category, cancer type, and anatomic site, can provide useful semantic supervision without paired slide--report data or curated caption corpora.  

Several limitations should be noted. ASTRA was not trained as a fully supervised segmentation model, and its localization maps are derived from tile--text similarity rather than direct boundary optimization. TCGA evaluation also relied on previously released tumor prediction maps from an external segmentation model rather than pathologist-delineated ground truth~\cite{skrede2026generalisation} ; the reported Dice values therefore reflect agreement with an external reference resource, not direct comparison with manual annotations. In addition, ASTRA requires offline extraction of embeddings from multiple foundation models, increasing preprocessing time and storage demands. Future work should evaluate ASTRA across broader institutions and rarer tumor types, ideally with larger manually annotated external cohorts, and test whether unified multi-foundation-model representations can support additional pathology tasks.

\vspace{6mm}

\noindent\textbf{\large{Conclusion}}\\

We introduced ASTRA, a pan-cancer pathology representation learning framework that unifies multiple pathology foundation-model embeddings and grounds slide representations in structured pathology annotation fields derived from slide-level metadata. Across diverse classification tasks and external localization evaluation, ASTRA showed that multi-foundation-model contextualization and lightweight semantic supervision can improve slide understanding without requiring paired reports or dense tumor labels. These findings support minimal structured pathology annotation fields as a scalable path toward semantically grounded pathology foundation models.

\vspace{6mm}

\noindent\textbf{\large{Methods}}\label{sec:method}\\

\noindent\textbf{Dataset and cohort design}\label{sec:dataset}\\
 
We use the Cooperative Human Tissue Network (CHTN) cohort, a multi-institutional repository of digitized hematoxylin-and-eosin-stained WSIs curated from six academic medical centers across the United States~\cite{chtn}, for all stages of ASTRA training and evaluation. Within this tissue bank, the cohort of $10{,}359$ WSIs is structurally organized into four primary classification categories: Malignant ($n=7{,}231$; 69.8\%), Normal Adjacent to Tumor ($n=2{,}043$; 19.7\%), Benign ($n=647$; 6.2\%), and Normal ($n=438$; 4.2\%). The malignant category is further grouped into three broad solid tumor types: carcinomas ($n=6{,}706$; 92.7\% of malignant cases), sarcomas ($n=371$; 5.1\%), and melanomas ($n=154$; 2.1\%). As illustrated in \textbf{Figure~\ref{fig:overview}A (left)}, the malignant cohort exhibits a highly diverse subtype distribution across 16 fine-grained cancer classes and a minor subset of uncategorized malignancies: squamous cell carcinoma ($n=1{,}294$), renal cell carcinoma ($n=1{,}112$), colorectal carcinoma ($n=806$), endometrial carcinoma ($n=690$), breast carcinoma ($n=574$), lung carcinoma ($n=529$), ovarian carcinoma ($n=374$), sarcoma group ($n=371$), thyroid carcinoma ($n=310$), neuroendocrine neoplasm ($n=222$), urothelial carcinoma ($n=197$), prostate carcinoma ($n=156$), melanoma ($n=154$), pancreatic carcinoma ($n=124$), hepatocellular carcinoma ($n=85$), cholangiocarcinoma ($n=37$), and an Others group comprising rare or unspecified malignancies ($n=196$). Each CHTN slide is associated with three structured clinical annotation fields routinely available in clinical practice: classification category, cancer type, and anatomic site. These labels are directly utilized to construct the descriptive slide-level pathology prompts for ASTRA (\textbf{Figure~\ref{fig:overview}A, right}).

The cohort was originally partitioned into an $80\%$ training split and a $20\%$ held-out test split by stratified random sampling on cancer class (seed~$=42$). All stages of ASTRA pretraining and slide--text alignment exclusively utilized the $80\%$ training split; the held-out test split remained strictly unseen throughout the entire pipeline and was not accessed during model training or selection. For downstream ABMIL classification, the $80\%$ training split was dynamically subdivided during each training run: $90\%$ of this split was used for model parameter updates and $10\%$ was reserved as a validation set for early stopping. Final performance was evaluated exclusively on the $20\%$ held-out test split and reported as mean~$\pm$~standard deviation over five independently seeded runs. No cross-validation across the full cohort was performed.

The three downstream classification tasks were defined on distinct subsets of the CHTN cohort. The 4-class classification-category task utilized the full cohort of $10{,}359$ slides, with a held-out test set of $n=2{,}072$. The 3-class major-group task (Carcinoma / Sarcoma / Melanoma) encompassed all $7{,}231$ malignant slides, with neuroendocrine Neoplasms integrated into the Carcinoma group, resulting in a held-out test set of $n=1{,}446$. The 16-class cancer-typing task was restricted to $7{,}035$ malignant slides belonging to the 16 predefined core subtypes (excluding a small number of rare tumors outside these panels), yielding a held-out test set of $n=1{,}405$.

Tumor localization was evaluated on a manually annotated subset of 380 CHTN slides drawn exclusively from the held-out test split and spanning all 16 cancer types: Breast carcinoma ($n=27$), Cholangiocarcinoma ($n=11$), Colorectal carcinoma ($n=21$), Endometrium carcinoma ($n=27$), Hepatocellular carcinoma ($n=26$), Lung carcinoma ($n=26$), Melanoma ($n=16$), Neuroendocrine neoplasms ($n=21$), Ovary carcinoma ($n=20$), Pancreas carcinoma ($n=17$), Prostate carcinoma ($n=20$), Renal cell carcinoma ($n=40$), Sarcoma group ($n=23$), Squamous cell carcinoma ($n=39$), Thyroid carcinoma ($n=29$), and Urothelial carcinoma ($n=17$). Tumor regions were delineated at the slide level to generate ground-truth masks. External generalization was assessed on an independent TCGA cohort comprising $1{,}686$ WSIs from four cancer types: LUAD ($n=476$), LUSC ($n=458$), PRAD ($n=365$), and BLCA ($n=387$). For TCGA evaluation, slide-specific prompts were instantiated from the available case-level cancer-type and anatomic-site metadata for these four classification categories, rather than from the full three-field annotation schema used in CHTN training. For example, a LUAD case was paired with the prompt \textit{``A histopathology whole-slide image of malignant lung adenocarcinoma from the lung''} We used the previously released slide-level tumor prediction maps from Skrede et al.~\cite{skrede2026generalisation} as the reference resource rather than manually annotated ground-truth masks. Because these reference masks were generated by an external segmentation model rather than by manual delineation, we excluded slides whose reference tumor masks covered less than 20\% of tissue area, in order to avoid unstable Dice estimates driven by extremely small predicted regions. No TCGA slides were used at any stage of training.

\noindent\textbf{Overview of ASTRA}\\
ASTRA is a pan-cancer representation learning framework that unifies heterogeneous pathology foundation-model embeddings and grounds them in structured pathology annotation fields. The full workflow proceeds from cohort-level annotation, to multi-foundation-model pretraining, to semantically aligned downstream inference. Starting from the CHTN cohort in \textbf{Figure~\ref{fig:overview}A}, each WSI is paired with three structured annotation fields: classification category, cancer class, and primary anatomic site, which are composed into slide-level pathology prompts. These prompts provide the semantic supervision used throughout the framework.

ASTRA pretraining then builds a shared representation space from multiple pathology foundation models (\textbf{Figure~\ref{fig:overview}B}). A WSI is first encoded by four pathology foundation models to form a pool of aligned tile embeddings. Local embedding crops are sampled from this shared pool, partially masked, and passed through a sparse MoE contextualizer, after which a decoder reconstructs masked embeddings across all four foundation-model spaces. In this way, ASTRA learns a unified contextual representation that preserves local spatial organization while remaining predictive across heterogeneous embedding spaces.

In the downstream application (\textbf{Figure~\ref{fig:overview}C}), contextualized tile embeddings are aggregated into a slide-level representation and aligned with structured pathology prompts through contrastive learning. The resulting representation supports two complementary applications: pan-cancer slide-level classification and zero-shot text-guided tumor localization. By reading out localization directly from tile--text similarity in the aligned representation space, ASTRA enables zero-shot tumor localization without requiring pixel-level labels, paired report data, or synthetic captions. 

\noindent\textbf{Shared spatial grid construction}\\
WSIs were processed using the TRIDENT pipeline~\cite{trident}. Tissue regions were identified by Otsu-based thresholding in HSV space and tessellated into non-overlapping tiles at $20\times$ magnification. Tile embeddings were extracted using four pathology foundation models at their native input resolutions: UNI~v2~\cite{chen2024uni} ($256\times256$ px, 1536-d), GigaPath~\cite{xu2024gigapath} ($256\times256$ px, 1536-d), CONCH~v1.5~\cite{lu2024conch} ($512\times512$ px, 768-d), and Virchow2~\cite{zimmermann2024virchow2} ($224\times224$ px, 2560-d).

Because these models operate at different tile sizes, receptive fields, and embedding dimensions, their raw feature maps are not directly comparable for dense spatial reasoning. All feature extraction was therefore anchored to a shared spatial grid with a fixed stride of 512 pixels, corresponding to a $25.6\,\mu$m step at $20\times$ magnification. This stride matches the largest native tile size among the four encoders (CONCH~v1.5, $512\times512$ px), ensuring that each grid coordinate maps to a single non-overlapping tissue region for every model without introducing sub-grid misalignment. For encoders with smaller native tile sizes, including UNI~v2 and GigaPath, multiple embeddings falling within the same $512\times512$ grid cell were average-pooled to produce a single aligned representation for that spatial location. Each valid tissue position is accordingly represented by four parallel embeddings, one per foundation model, defined on a common integer lattice.

\noindent\textbf{Tissue-aware local crop construction and asymmetric masking}\\
At each training step, we sample a $16\times16$ tile window from a slide at $20\times$ magnification. Under the shared 512-pixel grid stride, this corresponds to a region of approximately $8{,}192\times8{,}192$ pixels (about $4.1\times4.1$ mm), a mesoscopic scale that captures microenvironmental organization including tumor--stroma interfaces, glandular architecture, and immune-rich regions, consistent with the local context scale adopted in TITAN~\cite{ding2025multimodal}. Candidate windows are drawn at random and accepted only if at least 55\% of the 256 grid positions contain valid tissue tiles under the CONCH~v1.5 anchor model. This threshold excludes background-dominated windows while preserving exposure to sparse or transitional tissue patterns that are important for robust localization. If no candidate satisfies the threshold after repeated sampling, the highest-coverage window is retained as a fallback.

Within the selected window, 64 of the 256 tile positions are revealed to the encoder and the remaining 192 are masked, giving a masking ratio of 75\%. Histopathological tile embeddings exhibit strong short-range spatial redundancy; at lower masking ratios, masked positions can often be recovered by local interpolation alone. At 75\% masking, reconstruction requires integrating information across a broader spatial context and encourages the encoder to model tissue organization rather than local continuity.

At each training step, the encoder input foundation model is sampled uniformly at random so that no single embedding space dominates pretraining. Visible embeddings from the selected model are projected into a shared latent space through model-specific two-layer MLP heads, which normalize dimensional heterogeneity before contextual processing.

\noindent\textbf{Spatial contextualization with sparse MoE transformer}\\
The 64 visible tokens entering the encoder span only a fraction of the $16\times16$ crop and originate from a single randomly selected foundation model. Tissue within a single crop is often morphologically heterogeneous: tumor epithelium, reactive stroma, inflammatory infiltrate, and necrotic regions may coexist within the same local neighborhood, whereas a standard dense FFN applies the same transformation to every token. We therefore replace the feed-forward network in each encoder block with a sparse Mixture-of-Experts (MoE) layer~\cite{shazeer2017outrageously}, which routes each token to a learned subset of expert networks according to its content. This allows distinct tissue compartments to be processed by different expert pathways without requiring explicit compartment labels. These routing patterns are examined directly through mechanistic analysis of expert routing (\textbf{Figure~\ref{fig:expert_routing_vis}}).

\noindent\textbf{Multi-target reconstruction across heterogeneous embedding spaces}\\
Contextualized visible tokens are used to reconstruct masked tissue positions in all four foundation-model embedding spaces, regardless of which model provided the encoder input. This cross-space predictiveness is the core pretraining objective: the encoder representation must generalize beyond the sampled input space.

\textit{Hierarchical decoder.}
Reconstruction is performed by a lightweight cross-attention decoder inspired by Hi-End-MAE~\cite{hiendmae}. The decoder contains three blocks connected to different encoder depths: the first, second, and third decoder blocks receive their cross-attention context from the 2nd, 4th, and 6th encoder layers, respectively, after block-specific linear projections into the decoder space (\textbf{Figure~\ref{fig:overview}B, right}). This staged coupling exposes early decoder steps to local, low-level morphology embeddings while later steps receive increasingly contextualized representations, and prevents the encoder from concentrating all reconstructive information in its final layer. Spatial and morphological structure must therefore remain accessible at intermediate depths, which is consequential for downstream localization.

\textit{Multi-target reconstruction loss.}
Decoder outputs are projected independently to each foundation model's native embedding dimension through per-model output heads. Reconstruction is supervised by the mean cosine distance between predicted and ground-truth embeddings, averaged over all masked positions and all four models:
\begin{equation}
\mathcal{L}_{\mathrm{recon}} = \frac{1}{4}\sum_{k=1}^{4}
\frac{1}{|\mathcal{M}_k|}\sum_{i\in\mathcal{M}_k}
\!\left(1 -
\frac{\hat{\mathbf{y}}_{k,i}^\top\mathbf{y}_{k,i}}
{\|\hat{\mathbf{y}}_{k,i}\|_2\|\mathbf{y}_{k,i}\|_2}\right),
\end{equation}
where $\mathcal{M}_k$ denotes masked positions with a valid tissue embedding under model $k$. Cosine distance is preferred over mean-squared error because foundation-model embeddings encode morphological identity primarily through direction rather than magnitude.

\textit{Expert load balancing.}
To prevent routing collapse, a load-balancing regularizer~\cite{shazeer2017outrageously} penalizes unequal token allocation across experts. The full pretraining objective is
\begin{equation}
\mathcal{L} = \mathcal{L}_{\mathrm{recon}} + \lambda\,\mathcal{L}_{\mathrm{moe}},
\end{equation}
where $\lambda$ controls the strength of the expert load-balancing term (see Implementation Details).

\noindent\textbf{Semantic alignment via structured annotations}\\
Pretraining yields tile representations that retain spatial context and cross-model predictiveness but are not yet explicitly grounded in structured pathology annotation fields. To introduce such semantic grounding, ASTRA aligns slide-level representations with structured text derived from three annotation fields available in the curated slide-level metadata: classification category, cancer type, and primary anatomic site.

\textit{Structured prompt construction.}
The three annotation fields are composed into a natural-language prompt using four fixed templates matched to the diagnostic context:

\begin{tcolorbox}[
  enhanced, breakable,
  colback=promptbg, colframe=promptborder,
  boxrule=0.6pt, arc=4pt,
  left=6pt, right=6pt, top=4pt, bottom=4pt,
  title={\small\textbf{Pathology Prompt Templates}},
  fonttitle=\small\bfseries, coltitle=white,
  attach boxed title to top left={yshift=-2mm, xshift=6pt},
  boxed title style={colback=promptborder, arc=2pt, boxrule=0pt},
]
\small
\begin{tabular}{@{}p{0.18\linewidth}p{0.76\linewidth}@{}}
\textit{Malignant} &
  ``\textit{A histopathology whole-slide image of malignant
  \textbf{[cancer type]} from the \textbf{[anatomic site]}.}'' \\[4pt]
\textit{Norm. Adj.} &
  ``\textit{A histopathology whole-slide image showing normal adjacent
  tissue from the \textbf{[anatomic site]}.}'' \\[4pt]
\textit{Benign} &
  ``\textit{A histopathology whole-slide image of benign
  \textbf{[cancer type]} from the \textbf{[anatomic site]}.}'' \\[4pt]
\textit{Normal} &
  ``\textit{A histopathology whole-slide image of normal tissue
  from the \textbf{[anatomic site]}.}'' \\
\end{tabular}
\end{tcolorbox}

Slots are filled directly from the clinical record, with no expert curation, free-text generation, or paired report data. This differs from vision-language models such as TITAN~\cite{ding2025multimodal} and CONCH~\cite{lu2024conch}, which rely on large slide--report corpora that are frequently unavailable in routine practice. Prompts are encoded by the frozen CONCH~v1.5 text encoder, producing $L_2$-normalized 512-dimensional text embeddings.

\textit{Slide-level contrastive alignment.}
Tile embeddings from all valid tissue positions in a slide are extracted with the pretrained ASTRA encoder and aggregated into a single slide embedding by Gated Attention-based Multiple Instance Learning (Gated ABMIL)~\cite{ilse2018attention}. The gated attention mechanism assigns content-dependent weights to tiles, allowing diagnostically informative regions to drive the slide-level representation. The resulting embedding is projected to 512 dimensions to match the text embedding space.

Slide and text embeddings are each $L_2$-normalized. For a batch of $B$ slide--text pairs $\{(\mathbf{s}_i, \mathbf{t}_i)\}_{i=1}^{B}$, the slide-to-text loss is
\begin{equation}
\mathcal{L}_{\mathrm{s\to t}} = -\frac{1}{B}\sum_{i=1}^{B}
\log\frac{\exp(\mathbf{s}_i^\top\mathbf{t}_i\,/\,\tau)}
{\sum_{j=1}^{B}\exp(\mathbf{s}_i^\top\mathbf{t}_j\,/\,\tau)},
\end{equation}
and the final alignment objective is the symmetric average
$\mathcal{L}_{\mathrm{symCL}} = \tfrac{1}{2}[\mathcal{L}_{\mathrm{s\to t}} + \mathcal{L}_{\mathrm{t\to s}}]$,
where $\tau$ is the contrastive temperature.

\noindent\textbf{Downstream evaluation}\\
The learned ASTRA representation was evaluated in two settings: pan-cancer slide-level classification and text-guided tumor localization (\textbf{Figure~\ref{fig:overview}C}).

\textit{Pan-cancer slide-level classification.}
At inference time, tile embeddings from any of the four foundation-model backbones (UNI~v2, GigaPath, CONCH~v1.5, Virchow2) are projected through the corresponding input projector and passed through the shared sparse MoE encoder. This backbone-agnostic design allows a single pretrained ASTRA model to be evaluated across all four foundation models without retraining.

To disentangle the contributions of architecture and pretraining strategy, we evaluated five contextualization configurations per backbone, following the protocol of TICON~\cite{ticon}. \textbf{Raw} uses non-contextualized tile embeddings as ABMIL input and serves as the per-backbone baseline. \textbf{TICON (ISO)} and \textbf{ASTRA (ISO)} are single-FM variants pretrained on tile embeddings from the same backbone used at downstream evaluation time, following the original TICON protocol. \textbf{TICON} and \textbf{ASTRA} are the corresponding multi-FM variants, pretrained jointly on tile embeddings from all four foundation models. Here, \textbf{ASTRA} refers to the full model, combining joint multi-FM pretraining with the sparse MoE encoder. This design allows us to compare the effect of sparse expert routing against a dense FFN under matched ISO settings, while also examining the benefit of joint multi-FM pretraining relative to the corresponding ISO variants.

Contextualized tile embeddings were aggregated per slide with Gated ABMIL~\cite{ilse2018attention} and classified with a linear head. We evaluated three slide-level classification tasks of increasing granularity: 4-class classification-category prediction, 3-class major-group prediction (Carcinoma / Sarcoma / Melanoma, with neuroendocrine tumors integrated into the Carcinoma group) on malignant slides, and 16-class cancer-typing prediction on malignant slides. Performance was assessed by accuracy (Acc), balanced accuracy (B-Acc), macro-averaged one-vs-rest specificity (Sp$^*$), and macro-averaged one-vs-rest AUC across five independent random seeds.

\textit{Text-guided tumor localization.}
For a given slide, contextualized tile embeddings are projected from 1536 to 512 dimensions using the trained slide projection head from the semantic alignment stage and then $L_2$-normalized. This projection aligns the visual embedding space with the fixed 512-dimensional output space of the frozen CONCH~v1.5 text encoder. The slide-specific pathology prompt, constructed from classification category, cancer type, and anatomic site using the templates described above, is encoded by the frozen CONCH~v1.5 text encoder to produce a 512-dimensional text embedding. Per-tile cosine similarity is computed as
\begin{equation}
  s_i = \frac{\mathbf{f}_i^\top \mathbf{t}}
             {\|\mathbf{f}_i\|_2\|\mathbf{t}\|_2},
\end{equation}
and the resulting similarity heatmap is thresholded at a fixed value (see Implementation Details) to generate a binary tumor prediction mask without per-slide normalization or adaptive thresholding.

Localization performance was evaluated on the 380-slide annotated CHTN subset and on the external TCGA cohort without fine-tuning on either. For both cohorts, slide-specific prompts were instantiated from the available structured annotation fields. Dice similarity coefficients were computed on a per-slide basis against the corresponding reference masks and then summarized by cancer type and by cohort. For CHTN, the reference masks were manually delineated ground-truth tumor annotations. For TCGA, we used previously released slide-level tumor prediction maps from Skrede et al.~\cite{skrede2026generalisation} as the reference resource rather than manually annotated ground-truth masks. Because these TCGA reference masks were generated by an external segmentation model rather than by manual delineation, we excluded slides whose reference tumor masks covered less than 20\% of tissue area in order to avoid unstable Dice estimates driven by extremely small predicted regions. No localization labels were used at any stage of training; tumor localization was read out directly from the aligned visual--text representation space induced by slide-level diagnostic supervision.

\noindent\textbf{Implementation details}\label{sec:implementation}

ASTRA pretraining was conducted for 500 epochs on 8 NVIDIA A100 80~GB GPUs using PyTorch Distributed Data Parallel with the NCCL backend. Each slide contributed 20 independently sampled crops per epoch; with a per-GPU batch size of 128 and 8 GPUs, the effective batch size was 1{,}024. Optimization used AdamW~\cite{kingma2014adam} with learning rate $2\times10^{-4}$, weight decay 0.05, and momentum parameters $(\beta_1,\beta_2)=(0.9,0.999)$, with linear warmup over the first 2{,}000 steps followed by cosine annealing to zero. Gradient norms were clipped to 1.0 throughout. The load-balancing coefficient for Sparse MoE pretraining was set to $\lambda=0.01$.

The text alignment stage and all downstream experiments were run on a single NVIDIA A100 80~GB GPU. The text alignment stage used AdamW with learning rate $10^{-4}$, weight decay 0.05, cosine annealing for 50 epochs, and batch size 32. The ABMIL aggregator used a hidden dimension of 256, gated attention dropout of 0.25, and a 512-dimensional slide projection head to match the fixed output dimension of the CONCH~v1.5 text encoder. For slide--text contrastive alignment, the symmetric InfoNCE objective used a temperature of $\tau=0.1$. Incomplete final batches were discarded to ensure a consistent batch size for contrastive training.

Downstream ABMIL classifiers were trained using the Adam optimizer (learning rate $10^{-4}$, weight decay $10^{-5}$) and a cosine annealing scheduler for up to 30 epochs. Specifically, for each of the $K=5$ seeds, 10\% of the training split was randomly held out as a stratified internal validation set for early stopping based on macro-averaged recall (patience $= 7$), ensuring that the final model selected for test-set evaluation demonstrated the best generalization on unseen training data. For text-guided localization, a fixed cosine similarity threshold of $\tau_{\mathrm{loc}}=0.15$ was applied uniformly across all slides.

\clearpage
\noindent\textbf{Supplementary Materials}

\setcounter{table}{0}
\setcounter{figure}{0}

\renewcommand{\thetable}{\arabic{table}}
\renewcommand{\thefigure}{\arabic{figure}}
\renewcommand{\tablename}{Supplementary Table}
\renewcommand{\figurename}{Supplementary Figure}

\noindent\mbox{}\par
\vspace{2mm}

\newcommand{\suptabsetup}{%
  \renewcommand{\arraystretch}{1.05}%
  \setlength{\tabcolsep}{4pt}%
}

\begin{table}[H]
\centering
\suptabsetup
\small
\caption{\textbf{4-category} classification (Malignant / Normal Adjacent To tumor / Benign / Normal; total $n=10{,}359$ with an 80\%/20\% train/test split). 
Values are mean\,$\pm$\,std over $K=5$ seeds on the held-out test set.
\textbf{Acc}: accuracy; \textbf{B-Acc}: balanced accuracy (macro-averaged recall);
\textbf{Sp$^*$}: macro one-vs-rest specificity; \textbf{AUC}: macro one-vs-rest AUC.
Subscripts report absolute percentage-point change versus \textit{Raw}
(\textcolor{PineGreen}{$\uparrow$} / \textcolor{BrickRed}{$\downarrow$} / \textcolor{gray}{=}).
\textbf{ASTRA (ISO)}: single-FM pretraining; \textbf{ASTRA}: full multi-FM pretraining.
Bold indicates the highest AUC within each backbone.} 
\label{stab:cls_category4}
\begin{tabular}{ll|cccc}
\toprule
\textbf{Backbone} & \textbf{Method} & \textbf{Acc} & \textbf{B-Acc} & \textbf{Sp$^*$} & \textbf{AUC} \\
\midrule
\multirow{5}{*}{\textbf{CONCH v1.5}}
& Raw             & $92.4\pm0.6$ & $73.2\pm0.9$ & $97.4\pm0.1$ & $96.7\pm0.1$ \\
\cdashline{2-6}[0.4pt/1.5pt]
& TICON (ISO)     & $92.4\pm0.6$\ppz{+0.0} & $73.1\pm0.3$\ppr{-0.1} & $97.4\pm0.2$\ppz{+0.0} & $96.8\pm0.2$\ppg{+0.1} \\
& ASTRA (ISO)     & $93.0\pm0.8$\ppg{+0.6} & $76.8\pm1.3$\ppg{+3.6} & $97.6\pm0.2$\ppg{+0.2} & $\mathbf{97.4}\pm0.1$\ppg{+0.7} \\
\cdashline{2-6}[0.4pt/1.5pt]
& TICON           & $92.6\pm0.3$\ppg{+0.2} & $72.7\pm1.2$\ppr{-0.5} & $97.4\pm0.1$\ppz{+0.0} & $96.8\pm0.2$\ppg{+0.1} \\
& ASTRA           & $92.7\pm0.4$\ppg{+0.3} & $75.6\pm0.8$\ppg{+2.4} & $97.6\pm0.1$\ppg{+0.2} & $97.1\pm0.4$\ppg{+0.4} \\
\midrule
\multirow{5}{*}{\textbf{UNI v2}}
& Raw             & $92.9\pm0.4$ & $75.4\pm0.7$ & $97.6\pm0.2$ & $97.1\pm0.3$ \\
\cdashline{2-6}[0.4pt/1.5pt]
& TICON (ISO)     & $92.9\pm0.4$\ppz{+0.0} & $75.5\pm1.0$\ppg{+0.1} & $97.5\pm0.1$\ppr{-0.1} & $97.2\pm0.2$\ppg{+0.1} \\
& ASTRA (ISO)     & $93.1\pm0.5$\ppg{+0.2} & $76.2\pm1.5$\ppg{+0.8} & $97.7\pm0.3$\ppg{+0.1} & $97.4\pm0.2$\ppg{+0.3} \\
\cdashline{2-6}[0.4pt/1.5pt]
& TICON           & $93.0\pm0.2$\ppg{+0.1} & $75.3\pm1.1$\ppr{-0.1} & $97.6\pm0.2$\ppz{+0.0} & $97.2\pm0.3$\ppg{+0.1} \\
& ASTRA           & $93.3\pm0.4$\ppg{+0.4} & $76.6\pm0.5$\ppg{+1.2} & $97.8\pm0.1$\ppg{+0.2} & $\mathbf{97.6}\pm0.2$\ppg{+0.5} \\
\midrule
\multirow{5}{*}{\textbf{Virchow2}}
& Raw             & $93.0\pm0.2$ & $75.4\pm0.7$ & $97.6\pm0.1$ & $97.2\pm0.4$ \\
\cdashline{2-6}[0.4pt/1.5pt]
& TICON (ISO)     & $92.9\pm0.3$\ppr{-0.1} & $76.3\pm1.2$\ppg{+0.9} & $97.7\pm0.1$\ppg{+0.1} & $97.1\pm0.3$\ppr{-0.1} \\
& ASTRA (ISO)     & $93.3\pm0.7$\ppg{+0.3} & $76.6\pm1.6$\ppg{+1.2} & $97.4\pm0.3$\ppr{-0.2} & $97.6\pm0.3$\ppg{+0.4} \\
\cdashline{2-6}[0.4pt/1.5pt]
& TICON           & $93.1\pm0.3$\ppg{+0.1} & $75.8\pm1.0$\ppg{+0.4} & $97.5\pm0.1$\ppr{-0.1} & $97.4\pm0.3$\ppg{+0.2} \\
& ASTRA           & $93.5\pm0.3$\ppg{+0.5} & $77.0\pm1.1$\ppg{+1.6} & $97.8\pm0.0$\ppg{+0.2} & $\mathbf{97.8}\pm0.1$\ppg{+0.6} \\
\midrule
\multirow{5}{*}{\textbf{GigaPath}}
& Raw             & $91.8\pm0.6$ & $75.2\pm1.3$ & $97.1\pm0.1$ & $96.9\pm0.2$ \\
\cdashline{2-6}[0.4pt/1.5pt]
& TICON (ISO)     & $92.0\pm0.5$\ppg{+0.2} & $75.7\pm1.3$\ppg{+0.5} & $97.2\pm0.2$\ppg{+0.1} & $96.5\pm0.2$\ppr{-0.4} \\
& ASTRA (ISO)     & $92.6\pm0.7$\ppg{+0.8} & $75.7\pm1.1$\ppg{+0.5} & $97.4\pm0.3$\ppg{+0.3} & $\mathbf{97.0}\pm0.2$\ppg{+0.1} \\
\cdashline{2-6}[0.4pt/1.5pt]
& TICON           & $92.0\pm0.7$\ppg{+0.2} & $74.9\pm0.7$\ppr{-0.3} & $97.2\pm0.3$\ppg{+0.1} & $96.6\pm0.3$\ppr{-0.3} \\
& ASTRA           & $92.7\pm0.4$\ppg{+0.9} & $75.7\pm2.1$\ppg{+0.5} & $97.4\pm0.2$\ppg{+0.3} & $\mathbf{97.0}\pm0.3$\ppg{+0.1} \\
\bottomrule
\end{tabular}
\end{table}

\begin{table}[t]
\centering
\suptabsetup
\small
\caption{\textbf{3-class major-group} prediction (Carcinoma / Sarcoma / Melanoma; total $n=7{,}231$ with an 80\%/20\% train/test split). 
Same reporting convention as Supplementary Table~\ref{stab:cls_category4}.}
\label{stab:cls_major3}
\begin{tabular}{ll|cccc}
\toprule
\textbf{Backbone} & \textbf{Method} & \textbf{Acc} & \textbf{B-Acc} & \textbf{Sp$^*$} & \textbf{AUC} \\
\midrule
\multirow{5}{*}{\textbf{CONCH v1.5}}
& Raw             & $98.5\pm0.1$ & $92.9\pm1.4$ & $98.1\pm0.6$ & $98.9\pm0.2$ \\
\cdashline{2-6}[0.4pt/1.5pt]
& TICON (ISO)     & $98.6\pm0.3$\ppg{+0.1} & $93.2\pm1.6$\ppg{+0.3} & $97.8\pm0.6$\ppr{-0.3} & $99.0\pm0.3$\ppg{+0.1} \\
& ASTRA (ISO)     & $98.4\pm0.3$\ppr{-0.1} & $93.2\pm0.5$\ppg{+0.3} & $98.2\pm0.6$\ppg{+0.1} & $99.2\pm0.3$\ppg{+0.3} \\
\cdashline{2-6}[0.4pt/1.5pt]
& TICON           & $98.5\pm0.3$\ppz{+0.0} & $93.0\pm1.0$\ppg{+0.1} & $98.0\pm0.4$\ppr{-0.1} & $98.7\pm0.5$\ppr{-0.2} \\
& ASTRA           & $98.6\pm0.3$\ppg{+0.1} & $91.3\pm3.0$\ppr{-1.6} & $97.9\pm0.7$\ppr{-0.2} & $\mathbf{99.4}\pm0.2$\ppg{+0.5} \\
\midrule
\multirow{5}{*}{\textbf{UNI v2}}
& Raw             & $98.7\pm0.1$ & $92.4\pm1.2$ & $98.0\pm0.7$ & $98.8\pm0.4$ \\
\cdashline{2-6}[0.4pt/1.5pt]
& TICON (ISO)     & $98.8\pm0.1$\ppg{+0.1} & $92.7\pm1.4$\ppg{+0.3} & $97.9\pm0.6$\ppr{-0.1} & $99.0\pm0.3$\ppg{+0.2} \\
& ASTRA (ISO)     & $98.8\pm0.2$\ppg{+0.1} & $93.1\pm1.2$\ppg{+0.7} & $98.2\pm0.5$\ppg{+0.2} & $99.2\pm0.4$\ppg{+0.4} \\
\cdashline{2-6}[0.4pt/1.5pt]
& TICON           & $98.7\pm0.2$\ppz{+0.0} & $92.5\pm1.9$\ppg{+0.1} & $98.1\pm0.8$\ppg{+0.1} & $98.9\pm0.2$\ppg{+0.1} \\
& ASTRA           & $99.0\pm0.2$\ppg{+0.3} & $93.6\pm1.2$\ppg{+1.2} & $98.4\pm0.4$\ppg{+0.4} & $\mathbf{99.4}\pm0.1$\ppg{+0.6} \\
\midrule
\multirow{5}{*}{\textbf{Virchow2}}
& Raw             & $98.6\pm0.2$ & $92.5\pm0.5$ & $98.1\pm0.7$ & $99.3\pm0.2$ \\
\cdashline{2-6}[0.4pt/1.5pt]
& TICON (ISO)     & $98.7\pm0.3$\ppg{+0.1} & $92.2\pm2.9$\ppr{-0.3} & $98.4\pm0.4$\ppg{+0.3} & $\mathbf{99.5}\pm0.1$\ppg{+0.2} \\
& ASTRA (ISO)     & $98.8\pm0.2$\ppg{+0.2} & $93.6\pm0.9$\ppg{+1.1} & $98.6\pm0.6$\ppg{+0.5} & $99.6\pm0.3$\ppg{+0.3} \\
\cdashline{2-6}[0.4pt/1.5pt]
& TICON           & $98.7\pm0.4$\ppg{+0.1} & $93.0\pm1.8$\ppg{+0.5} & $98.3\pm0.6$\ppg{+0.2} & $99.4\pm0.3$\ppg{+0.1} \\
& ASTRA           & $98.9\pm0.1$\ppg{+0.3} & $94.7\pm0.7$\ppg{+2.2} & $98.7\pm0.5$\ppg{+0.6} & $\mathbf{99.7}\pm0.3$\ppg{+0.4} \\
\midrule
\multirow{5}{*}{\textbf{GigaPath}}
& Raw             & $98.2\pm0.2$ & $91.7\pm2.1$ & $97.0\pm0.7$ & $99.1\pm0.4$ \\
\cdashline{2-6}[0.4pt/1.5pt]
& TICON (ISO)     & $97.8\pm0.2$\ppr{-0.4} & $90.7\pm1.1$\ppr{-1.0} & $96.5\pm0.6$\ppr{-0.5} & $99.3\pm0.1$\ppg{+0.2} \\
& ASTRA (ISO)     & $98.5\pm0.3$\ppg{+0.3} & $92.5\pm2.4$\ppg{+0.8} & $97.6\pm0.7$\ppg{+0.6} & $99.1\pm0.3$\ppz{+0.0} \\
\cdashline{2-6}[0.4pt/1.5pt]
& TICON           & $98.0\pm0.2$\ppr{-0.2} & $88.9\pm2.6$\ppr{-2.8} & $96.4\pm0.9$\ppr{-0.6} & $99.1\pm0.6$\ppz{+0.0} \\
& ASTRA           & $98.6\pm0.1$\ppg{+0.4} & $93.4\pm1.6$\ppg{+1.7} & $98.1\pm0.7$\ppg{+1.1} & $\mathbf{99.4}\pm0.1$\ppg{+0.3} \\
\bottomrule
\end{tabular}
\end{table}

\begin{table}[t]
\centering
\suptabsetup
\small
\caption{\textbf{16-class cancer typing} classification (malignant slides only; total $n=7{,}035$ with an 80\%/20\% train/test split). 
Same reporting convention as Supplementary Table~\ref{stab:cls_category4}.}
\label{stab:cls_cancer16}
\begin{tabular}{ll|cccc}
\toprule
\textbf{Backbone} & \textbf{Method} & \textbf{Acc} & \textbf{B-Acc} & \textbf{Sp$^*$} & \textbf{AUC} \\
\midrule
\multirow{5}{*}{\textbf{CONCH v1.5}}
& Raw             & $90.5\pm0.3$ & $86.1\pm0.7$ & $99.3\pm0.0$ & $98.9\pm0.1$ \\
\cdashline{2-6}[0.4pt/1.5pt]
& TICON (ISO)     & $90.3\pm0.3$\ppr{-0.2} & $84.9\pm0.5$\ppr{-1.2} & $99.3\pm0.0$\ppz{+0.0} & $98.9\pm0.1$\ppz{+0.0} \\
& ASTRA (ISO)     & $90.9\pm0.2$\ppg{+0.4} & $86.9\pm0.7$\ppg{+0.8} & $99.3\pm0.0$\ppz{+0.0} & $99.1\pm0.2$\ppg{+0.2} \\
\cdashline{2-6}[0.4pt/1.5pt]
& TICON           & $90.1\pm0.3$\ppr{-0.4} & $86.4\pm0.6$\ppg{+0.3} & $99.3\pm0.0$\ppz{+0.0} & $99.0\pm0.1$\ppg{+0.1} \\
& ASTRA           & $91.2\pm0.4$\ppg{+0.7} & $87.2\pm0.9$\ppg{+1.1} & $99.4\pm0.0$\ppg{+0.1} & $\mathbf{99.2}\pm0.1$\ppg{+0.3} \\
\midrule
\multirow{5}{*}{\textbf{UNI v2}}
& Raw             & $90.7\pm0.2$ & $85.6\pm0.8$ & $99.4\pm0.0$ & $98.9\pm0.1$ \\
\cdashline{2-6}[0.4pt/1.5pt]
& TICON (ISO)     & $90.4\pm0.4$\ppr{-0.3} & $85.1\pm0.9$\ppr{-0.5} & $99.3\pm0.0$\ppz{+0.0} & $98.9\pm0.1$\ppz{+0.0} \\
& ASTRA (ISO)     & $91.4\pm0.5$\ppg{+0.7} & $86.5\pm1.1$\ppg{+0.9} & $99.5\pm0.0$\ppg{+0.1} & $99.1\pm0.1$\ppg{+0.2} \\
\cdashline{2-6}[0.4pt/1.5pt]
& TICON           & $89.9\pm0.5$\ppr{-0.8} & $84.0\pm0.4$\ppr{-1.6} & $99.3\pm0.0$\ppr{-0.1} & $98.8\pm0.1$\ppr{-0.1} \\
& ASTRA           & $91.5\pm0.3$\ppg{+0.8} & $86.8\pm0.5$\ppg{+1.2} & $99.5\pm0.0$\ppg{+0.1} & $\mathbf{99.2}\pm0.1$\ppg{+0.3} \\
\midrule
\multirow{5}{*}{\textbf{Virchow2}}
& Raw             & $90.9\pm0.4$ & $85.0\pm0.5$ & $99.4\pm0.0$ & $98.8\pm0.2$ \\
\cdashline{2-6}[0.4pt/1.5pt]
& TICON (ISO)     & $90.6\pm0.5$\ppr{-0.4} & $84.2\pm1.1$\ppr{-0.7} & $99.4\pm0.0$\ppz{+0.0} & $99.0\pm0.1$\ppg{+0.2} \\
& ASTRA (ISO)     & $91.6\pm0.3$\ppg{+0.7} & $85.7\pm0.4$\ppg{+0.7} & $99.5\pm0.0$\ppg{+0.1} & $99.0\pm0.1$\ppg{+0.2} \\
\cdashline{2-6}[0.4pt/1.5pt]
& TICON           & $89.9\pm0.6$\ppr{-1.1} & $83.9\pm0.5$\ppr{-1.1} & $99.3\pm0.0$\ppr{-0.1} & $98.8\pm0.3$\ppz{+0.0} \\
& ASTRA           & $91.8\pm0.5$\ppg{+0.9} & $86.2\pm0.7$\ppg{+1.2} & $99.5\pm0.0$\ppg{+0.1} & $\mathbf{99.1}\pm0.3$\ppg{+0.3} \\
\midrule
\multirow{5}{*}{\textbf{GigaPath}}
& Raw             & $87.9\pm0.5$ & $81.3\pm2.7$ & $99.2\pm0.0$ & $98.5\pm0.1$ \\
\cdashline{2-6}[0.4pt/1.5pt]
& TICON (ISO)     & $87.8\pm0.4$\ppz{+0.0} & $82.0\pm0.9$\ppg{+0.7} & $99.2\pm0.0$\ppz{+0.0} & $98.5\pm0.2$\ppr{-0.1} \\
& ASTRA (ISO)     & $90.4\pm0.3$\ppg{+2.6} & $84.6\pm0.9$\ppg{+3.3} & $99.3\pm0.0$\ppg{+0.2} & $98.9\pm0.1$\ppg{+0.4} \\
\cdashline{2-6}[0.4pt/1.5pt]
& TICON           & $87.5\pm0.3$\ppr{-0.4} & $80.5\pm0.8$\ppr{-0.8} & $99.1\pm0.0$\ppz{+0.0} & $98.5\pm0.1$\ppz{+0.0} \\
& ASTRA           & $90.7\pm0.2$\ppg{+2.9} & $85.3\pm0.5$\ppg{+4.0} & $99.4\pm0.0$\ppg{+0.2} & $\mathbf{99.0}\pm0.1$\ppg{+0.4} \\
\bottomrule
\end{tabular}
\end{table}

\begin{table}[t]
\centering
\suptabsetup
\small
\caption{\textbf{Text-guided tumor localization on the CHTN annotated subset} ($n=380$), stratified by cancer type. Values are mean Dice~$\pm$~SD and median Dice across slides. \textit{Overall} summarizes all 380 cases; \textit{Macro} denotes the unweighted mean across the 16 cancer types. Localization results are reported for ASTRA only, as this evaluation relies on semantic grounding from routine structured pathology annotation fields and is not directly comparable to models pretrained under different supervision or alignment settings.}
\label{stab:chtn_dice}
\begin{tabular}{lrrr}
\toprule
\textbf{Cancer type} & $n$ & \textbf{Mean Dice $\pm$ SD} & \textbf{Median Dice} \\
\midrule
Breast Carcinoma                  & 27  & $0.879 \pm 0.222$ & $0.963$ \\
Cholangiocarcinoma                & 11  & $0.912 \pm 0.111$ & $0.947$ \\
Colorectal Adenocarcinoma         & 21  & $0.877 \pm 0.107$ & $0.904$ \\
Endometrial Carcinoma             & 27  & $0.864 \pm 0.217$ & $0.973$ \\
Hepatocellular Carcinoma          & 26  & $0.960 \pm 0.085$ & $0.985$ \\
Lung Carcinoma                    & 26  & $0.839 \pm 0.221$ & $0.910$ \\
Melanoma                          & 16  & $0.894 \pm 0.167$ & $0.965$ \\
Neuroendocrine Neoplasm           & 21  & $0.943 \pm 0.119$ & $0.980$ \\
Ovarian Carcinoma                 & 20  & $0.892 \pm 0.154$ & $0.962$ \\
Pancreatic Adenocarcinoma         & 17  & $0.818 \pm 0.184$ & $0.886$ \\
Prostatic Adenocarcinoma          & 20  & $0.913 \pm 0.094$ & $0.952$ \\
Renal Cell Carcinoma              & 40  & $0.937 \pm 0.125$ & $0.979$ \\
Sarcoma                           & 23  & $0.955 \pm 0.106$ & $0.991$ \\
Squamous Cell Carcinoma           & 39  & $0.807 \pm 0.243$ & $0.964$ \\
Thyroid Carcinoma                 & 29  & $0.963 \pm 0.029$ & $0.972$ \\
Urothelial Carcinoma              & 17  & $0.926 \pm 0.131$ & $0.981$ \\
\midrule
\textit{Overall}                  & 380 & $0.897 \pm 0.167$ & $0.969$ \\
\textit{Macro}                    & --  & $0.899$           & --      \\
\bottomrule
\end{tabular}
\end{table}

\begin{table}[t]
\centering
\suptabsetup
\small
\caption{\textbf{Text-guided tumor localization on the external TCGA cohort} ($n=1{,}686$), stratified by cancer type.
The CHTN-trained model and fixed threshold were applied directly to TCGA in a zero-shot setting, without dataset-specific adjustment or fine-tuning.
\textit{Overall} summarizes all evaluated cases; \textit{Macro} denotes the unweighted mean across the four cancer types.}
\label{stab:tcga_dice}
\begin{tabular}{lrrr}
\toprule
\textbf{Cancer type} & $n$ & \textbf{Mean Dice $\pm$ SD} & \textbf{Median Dice} \\
\midrule
Lung adenocarcinoma (LUAD)          & 476  & $0.779 \pm 0.147$ & $0.819$ \\
Lung squamous cell carcinoma (LUSC) & 458  & $0.817 \pm 0.131$ & $0.858$ \\
Prostate adenocarcinoma (PRAD)      & 365  & $0.628 \pm 0.219$ & $0.656$ \\
Urothelial carcinoma (BLCA)         & 387  & $0.697 \pm 0.194$ & $0.737$ \\
\midrule
\textit{Overall}                    & 1686 & $0.738 \pm 0.187$ & $0.796$ \\
\textit{Macro}                      & --   & $0.730$           & --      \\
\bottomrule
\end{tabular}
\end{table}

\clearpage
\begin{center}
\includegraphics[width=\textwidth]{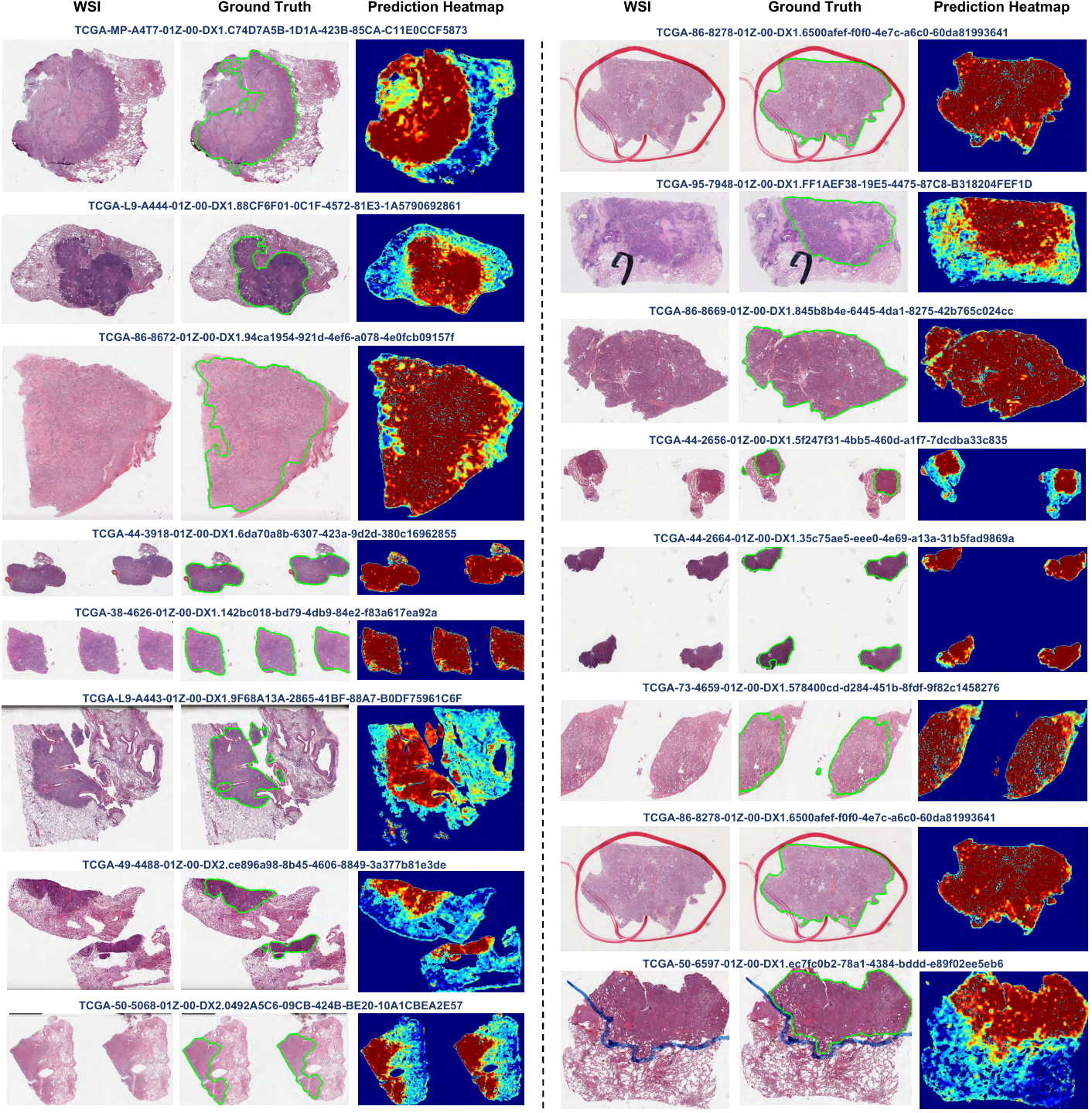}
\end{center}
\captionof{figure}{\textbf{Representative ASTRA text-guided tumor localization in TCGA lung adenocarcinoma (LUAD).}
Columns show the H\&E WSI thumbnail, the reference
tumor contour overlaid in green, and the prediction heatmap generated
from tile-text cosine similarity between ASTRA tile embeddings and the
slide-specific pathology prompt. Warmer colors indicate higher similarity.
TCGA slide identifiers are shown above each example.}
\label{sfig:tcga_luad}

\vspace{3mm}

\begin{center}
\includegraphics[width=\textwidth]{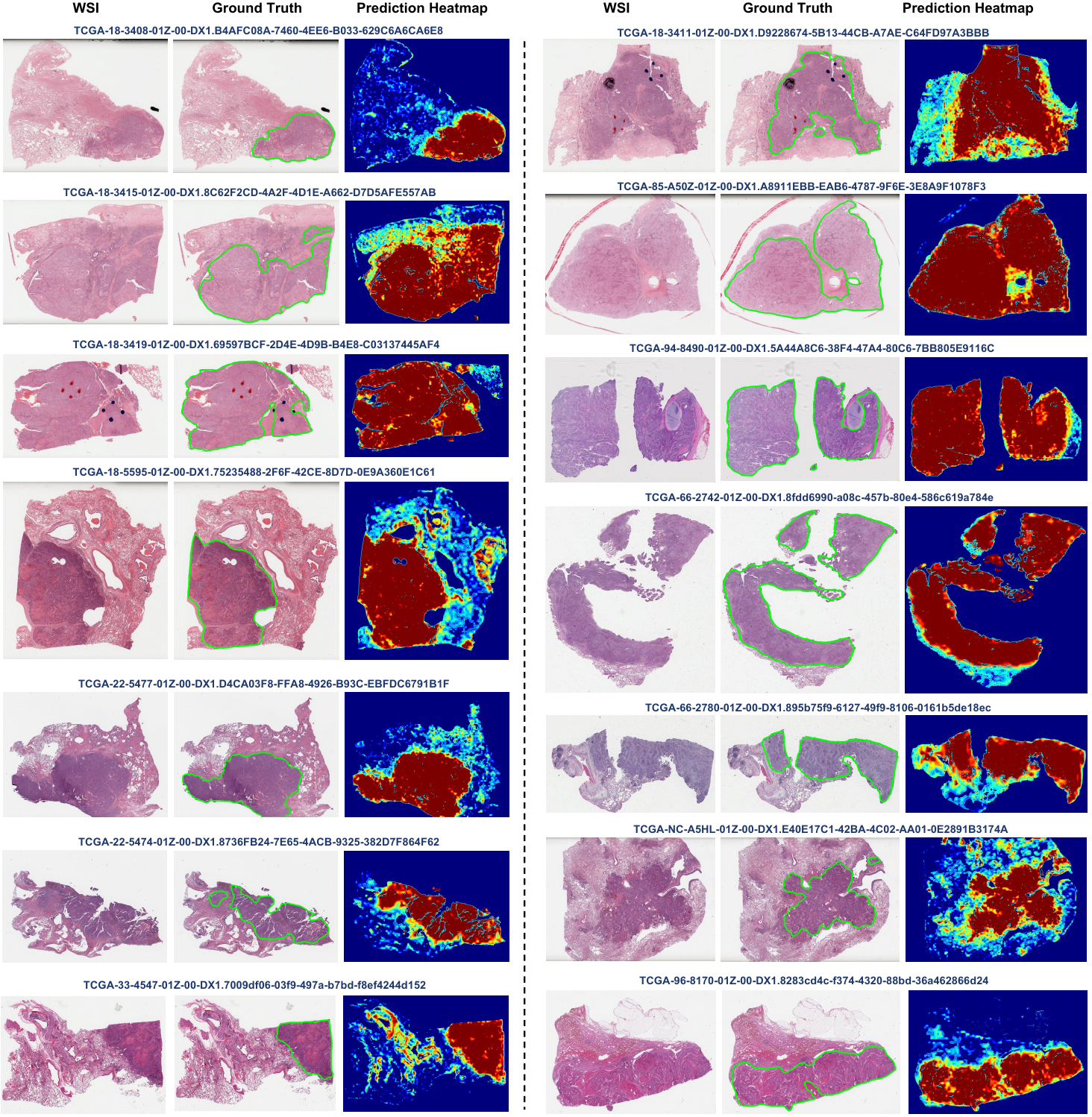}
\end{center}
\captionof{figure}{\textbf{Representative ASTRA text-guided tumor localization in TCGA lung squamous cell carcinoma (LUSC).}
Columns show the H\&E WSI thumbnail, the reference
tumor contour overlaid in green, and the prediction heatmap generated
from tile-text cosine similarity between ASTRA tile embeddings and the
slide-specific pathology prompt. Warmer colors indicate higher similarity.
TCGA slide identifiers are shown above each example.}
\label{sfig:tcga_lusc}


\newpage
\begin{center}
\includegraphics[width=\textwidth]{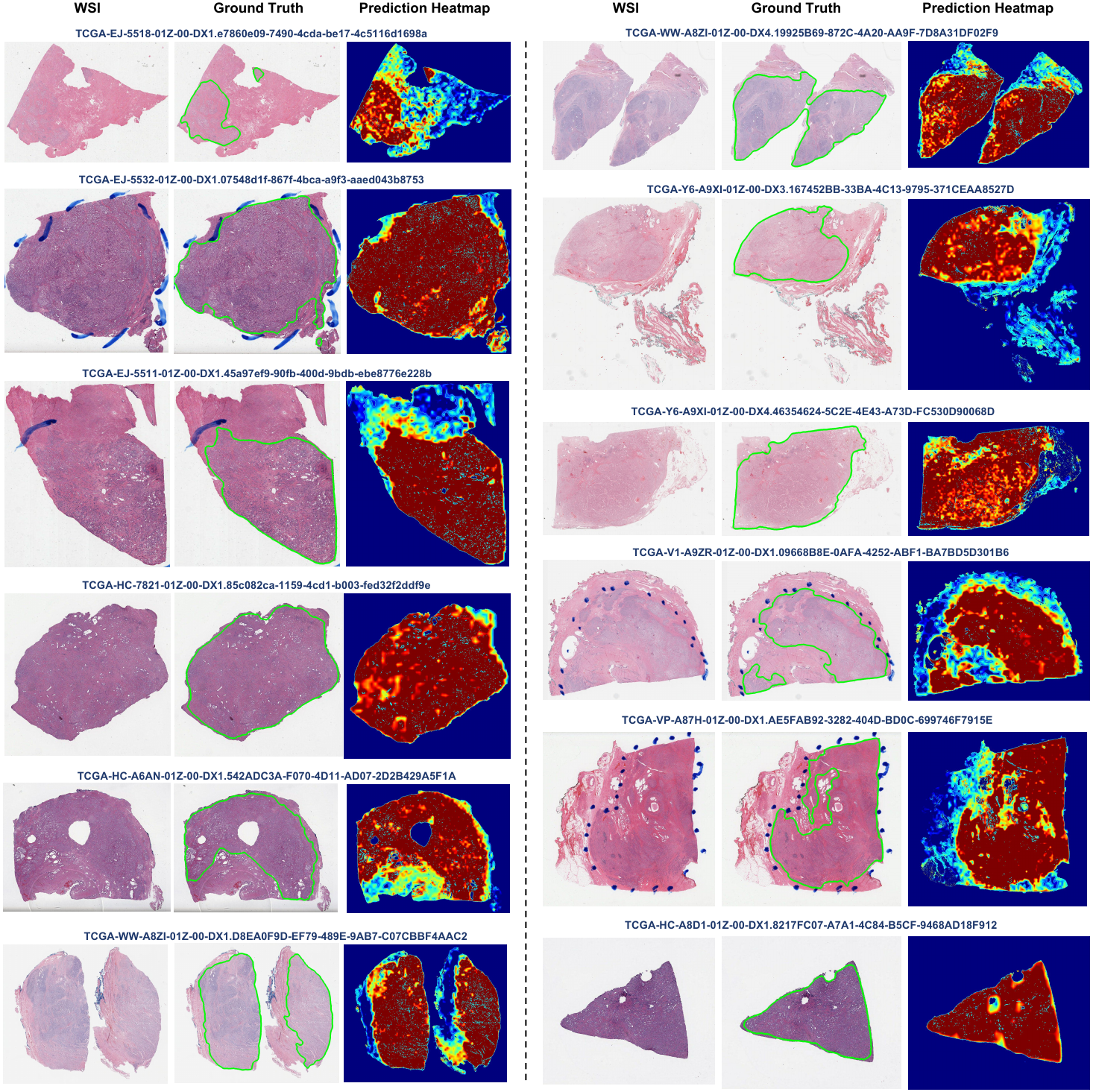}
\end{center}
\captionof{figure}{\textbf{Representative ASTRA text-guided tumor localization in TCGA prostate adenocarcinoma (PRAD).}
Columns show the H\&E WSI thumbnail, the reference
tumor contour overlaid in green, and the prediction heatmap generated
from tile-text cosine similarity between ASTRA tile embeddings and the
slide-specific pathology prompt. Warmer colors indicate higher similarity.
TCGA slide identifiers are shown above each example.}
\label{sfig:tcga_prad}

\begin{center}
\includegraphics[width=\textwidth]{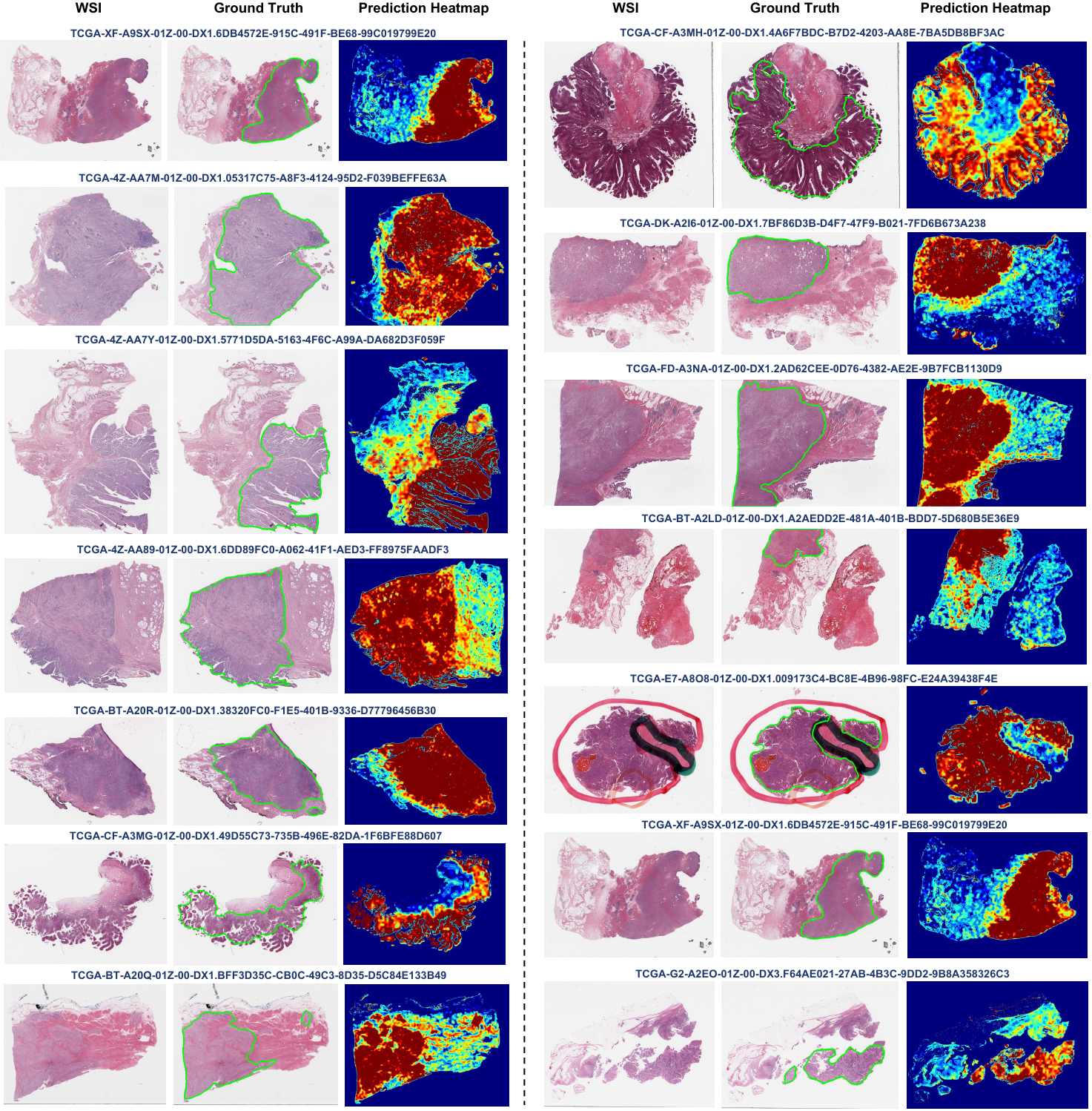}
\end{center}
\captionof{figure}{\textbf{Representative ASTRA text-guided tumor localization in TCGA bladder urothelial carcinoma (BLCA).}
Columns show the H\&E WSI thumbnail, the reference
tumor contour overlaid in green, and the prediction heatmap generated
from tile-text cosine similarity between ASTRA tile embeddings and the
slide-specific pathology prompt. Warmer colors indicate higher similarity.
TCGA slide identifiers are shown above each example.}
\label{sfig:tcga_blca}

\textbf{Code and data availability}
The processed data and underlying code for this study will be made available upon reasonable request to the
corresponding author. The Cooperative Human Tissue Network (CHTN) cohort~\cite{chtn} is subject
to institutional data access restrictions and is available through the
Cooperative Human Tissue Network under appropriate approvals and access
agreements. TCGA  WSIs can be accessed through the TCGA
Research Network (\url{https://www.cancer.gov/tcga}). The TCGA tumor
segmentation predictions used for localization evaluation in BLCA, LUAD,
LUSC, and PRAD are publicly available at Zenodo
(\url{https://zenodo.org/records/18481478}).

\noindent\textbf{Author Contributions}

T.W. conceptualized and designed the study; developed the overall framework and architectural design; implemented the methodology; conducted the experiments; analyzed the results; created the visualizations and figures; and wrote the manuscript.
Z.S. contributed to model design and development; assisted with experiments and analysis; and contributed to writing and revising the manuscript. A.R.A. and U.S. assisted with methodology implementation, experiments, and data analysis; and reviewed and revised the manuscript. L.G. and C.R. performed the histopathology annotations and contributed to data curation.
W.C. and A.P. provided clinical insight; approved the study design; evaluated the results as an expert attending pathologist; and reviewed and revised the manuscript. M.K.K.N. conceptualized and designed the study; supervised the research; validated the methodology and results; provided funding support; and edited and revised the manuscript.

\textbf{Competing Interests}
The authors declare no competing interests.

\noindent\textbf{Acknowledgments}

We thank the patients who contributed tissue samples to the Cooperative Human Tissue Network (CHTN) cohort and acknowledge the Cooperative Human Tissue Network for providing access to the data used in this study. We gratefully thank Dr.~Skrede for making the TCGA tumor segmentation predictions publicly available and enabling their use in our external localization evaluation. This work used the Ohio Supercomputer Center's high-performance computing resources through its collaboration with The Ohio State University College of Medicine. We also thank the Department of Pathology and the Comprehensive Cancer Center at The Ohio State University for their support.

\noindent\textbf{Funding}

This project was supported in part by R01 CA276301 (PIs: Niazi and Chen) from the National Cancer Institute, Pelotonia under IRP CC13702 (PIs: Niazi, Vilgelm, and Roy), and by the Department of Pathology and the Comprehensive Cancer Center at The Ohio State University. The content is solely the responsibility of the authors and does not necessarily represent the official views of the National Cancer Institute, the National Institutes of Health, or The Ohio State University.

\noindent\textbf{Ethics Approval and Consent to Participate}

This study involved secondary analysis of retrospective, fully de-identified clinical and histopathology data obtained from existing institutional and public repositories. Under applicable regulations, the use of de-identified data does not constitute human subjects research. Therefore, Institutional Review Board (IRB) approval was not required, and informed consent to participate was waived.

\end{spacing}
\end{spacing}
\begin{nolinenumbers}
\clearpage

\noindent\textbf{\large{References}}

\begin{spacing}{0.9}
\bibliographystyle{naturemag}
\bibliography{references}

@article{chen2024uni,
  title={Towards a General-Purpose Foundation Model for Computational Pathology},
  author={Chen, Richard J and Ding, Tong and Lu, Ming Y and Williamson, Drew FK 
          and Jaume, Guillaume and Chen, Bowen and Zhang, Andrew and Shao, Daniel 
          and Song, Andrew H and Shaban, Muhammad and others},
  journal={Nature Medicine},
  publisher={Nature Publishing Group},
  year={2024}
}

@article{su2025streamline,
  title={Streamline pathology foundation model by cross-magnification distillation},
  author={Su, Ziyu and Akbar, Abdul Rehman and Sajjad, Usama and Parwani, Anil V and Niazi, Muhammad Khalid Khan},
  journal={arXiv preprint arXiv:2509.23097},
  year={2025}
}

@article{skrede2026generalisation,
  title={Generalisation of automatic tumour segmentation in histopathological whole-slide images across multiple cancer types},
  author={Skrede, Ole-Johan and Pradhan, Manohar and Isaksen, Maria Xepapadakis and Hveem, Tarjei Sveinsgjerd and Vlatkovic, Ljiljana and Nesbakken, Arild and Lindemann, Kristina and Kristensen, Gunnar B and Kasius, Jenneke and Zeimet, Alain G and others},
  journal={npj Precision Oncology},
  year={2026},
  publisher={Nature Publishing Group UK London}
}

@article{lu2024conch,
  title={A visual-language foundation model for computational pathology},
  author={Lu, Ming Y and Chen, Bowen and Williamson, Drew FK and Chen, Richard J 
          and Liang, Ivy and Ding, Tong and Jaume, Guillaume and Odintsov, Igor 
          and Le, Long Phi and Gerber, Georg and others},
  journal={Nature Medicine},
  pages={863--874},
  volume={30},
  year={2024},
  publisher={Nature Publishing Group}
}

@article{zimmermann2024virchow2,
  title={Virchow2: Scaling Self-Supervised Mixed Magnification Models in Pathology},
  author={Zimmermann, Eric and Vorontsov, Eugene and Viret, Julian and Casson, Adam 
          and Zelechowski, Michal and Shaikovski, George and Tenenholtz, Neil 
          and Hall, James and Fuchs, Thomas and Fusi, Nicolo and Liu, Siqi 
          and Severson, Kristen},
  journal={arXiv preprint arXiv:2408.00738},
  year={2024}
}

@article{xu2024gigapath,
  title={A whole-slide foundation model for digital pathology from real-world data},
  author={Xu, Hanwen and Usuyama, Naoto and Bagga, Jaspreet and Zhang, Sheng 
          and Rao, Rajesh and Naumann, Tristan and Wong, Cliff and Gero, Zelalem 
          and Gonz{\'a}lez, Javier and Gu, Yu and others},
  journal={Nature},
  year={2024},
  publisher={Nature Publishing Group UK London}
}

@article{shao2021transmil,
  title={Transmil: Transformer based correlated multiple instance learning for whole slide image classification},
  author={Shao, Zhuchen and Bian, Hao and Chen, Yang and Wang, Yifeng and Zhang, Jian and Ji, Xiangyang and others},
  journal={Advances in neural information processing systems},
  volume={34},
  pages={2136--2147},
  year={2021}
}

@article{lu2021data,
  title={Data-efficient and weakly supervised computational pathology on whole-slide images},
  author={Lu, Ming Y and Williamson, Drew FK and Chen, Tiffany Y and Chen, Richard J and Barbieri, Matteo and Mahmood, Faisal},
  journal={Nature biomedical engineering},
  volume={5},
  number={6},
  pages={555--570},
  year={2021},
  publisher={Nature Publishing Group UK London}
}

@article{ding2025multimodal,
  title={A multimodal whole-slide foundation model for pathology},
  author={Ding, Tong and Wagner, Sophia J and Song, Andrew H and Chen, Richard J and Lu, Ming Y and Zhang, Andrew and Vaidya, Anurag J and Jaume, Guillaume and Shaban, Muhammad and Kim, Ahrong and others},
  journal={Nature medicine},
  pages={1--13},
  year={2025},
  publisher={Nature Publishing Group US New York}
}

@article{trident,
  title={Accelerating Data Processing and Benchmarking of AI Models for Pathology},
  author={Zhang, Andrew and Jaume, Guillaume and Vaidya, Anurag and Ding, Tong and Mahmood, Faisal},
  journal={arXiv preprint arXiv:2502.06750},
  year={2025}
}

@article{shazeer2017outrageously,
  title={Outrageously large neural networks: The sparsely-gated mixture-of-experts layer},
  author={Shazeer, Noam and Mirhoseini, Azalia and Maziarz, Krzysztof and Davis, Andy and Le, Quoc and Hinton, Geoffrey and Dean, Jeff},
  journal={arXiv preprint arXiv:1701.06538},
  year={2017}
}

@article{hiendmae,
  title={Hi-End-MAE: Hierarchical encoder-driven masked autoencoders are stronger vision learners for medical image segmentation},
  author={Tang, Fenghe and Yao, Qingsong and Ma, Wenxin and Wu, Chenxu and Jiang, Zihang and Zhou, S Kevin},
  journal={Medical Image Analysis},
  pages={103770},
  year={2025},
  publisher={Elsevier}
}

@article{baxi2022digital,
  title={Digital pathology and artificial intelligence in translational medicine and clinical practice},
  author={Baxi, Vipul and Edwards, Robin and Montalto, Michael and Saha, Saurabh},
  journal={Modern Pathology},
  volume={35},
  number={1},
  pages={23--32},
  year={2022},
  publisher={Elsevier}
}

@article{niazi2019digital,
  title={Digital pathology and artificial intelligence},
  author={Niazi, Muhammad Khalid Khan and Parwani, Anil V and Gurcan, Metin N},
  journal={The lancet oncology},
  volume={20},
  number={5},
  pages={e253--e261},
  year={2019},
  publisher={Elsevier}
}

@article{rajpurkar2022ai,
  title={AI in health and medicine},
  author={Rajpurkar, Pranav and Chen, Emma and Banerjee, Oishi and Topol, Eric J},
  journal={Nature medicine},
  volume={28},
  number={1},
  pages={31--38},
  year={2022},
  publisher={Nature Publishing Group US New York}
}

@article{akbar2025learning,
  title={Learning the Language of Histopathology Images reveals Prognostic Subgroups in Invasive Lung Adenocarcinoma Patients},
  author={Akbar, Abdul Rehman and Sajjad, Usama and Su, Ziyu and Li, Wencheng and Xing, Fei and Ruiz, Jimmy and Chen, Wei and Niazi, Muhammad Khalid Khan},
  journal={arXiv preprint arXiv:2508.16742},
  year={2025}
}

@inproceedings{ronneberger2015u,
  title={U-net: Convolutional networks for biomedical image segmentation},
  author={Ronneberger, Olaf and Fischer, Philipp and Brox, Thomas},
  booktitle={International Conference on Medical image computing and computer-assisted intervention},
  pages={234--241},
  year={2015},
  organization={Springer}
}

@article{isensee2021nnu,
  title={nnU-Net: a self-configuring method for deep learning-based biomedical image segmentation},
  author={Isensee, Fabian and Jaeger, Paul F and Kohl, Simon AA and Petersen, Jens and Maier-Hein, Klaus H},
  journal={Nature methods},
  volume={18},
  number={2},
  pages={203--211},
  year={2021},
  publisher={Nature Publishing Group US New York}
}

@article{van2021hooknet,
  title={HookNet: Multi-resolution convolutional neural networks for semantic segmentation in histopathology whole-slide images},
  author={Van Rijthoven, Mart and Balkenhol, Maschenka and Sili{\c{n}}a, Karina and Van Der Laak, Jeroen and Ciompi, Francesco},
  journal={Medical image analysis},
  volume={68},
  pages={101890},
  year={2021},
  publisher={Elsevier}
}

@article{choi20212020,
  title={The 2020 WHO classification of tumors of soft tissue: selected changes and new entities},
  author={Choi, Joon Hyuk and Ro, Jae Y},
  journal={Advances in anatomic pathology},
  volume={28},
  number={1},
  pages={44--58},
  year={2021},
  publisher={LWW}
}

@article{wang2022label,
  title={Label cleaning multiple instance learning: Refining coarse annotations on single whole-slide images},
  author={Wang, Zhenzhen and Saoud, Carla and Wangsiricharoen, Sintawat and James, Aaron W and Popel, Aleksander S and Sulam, Jeremias},
  journal={IEEE transactions on medical imaging},
  volume={41},
  number={12},
  pages={3952--3968},
  year={2022},
  publisher={IEEE}
}

@article{lu2023foundational,
  title={A foundational multimodal vision language AI assistant for human pathology},
  author={Lu, Ming Y and Chen, Bowen and Williamson, Drew FK and Chen, Richard J and Ikamura, Kenji and Gerber, Georg and Liang, Ivy and Le, Long Phi and Ding, Tong and Parwani, Anil V and others},
  journal={arXiv preprint arXiv:2312.07814},
  year={2023}
}

@article{verghese2023computational,
  title={Computational pathology in cancer diagnosis, prognosis, and prediction--present day and prospects},
  author={Verghese, Gregory and Lennerz, Jochen K and Ruta, Danny and Ng, Wen and Thavaraj, Selvam and Siziopikou, Kalliopi P and Naidoo, Threnesan and Rane, Swapnil and Salgado, Roberto and Pinder, Sarah E and others},
  journal={The Journal of pathology},
  volume={260},
  number={5},
  pages={551--563},
  year={2023},
  publisher={Wiley Online Library}
}

@article{ticon,
  title={TICON: A Slide-Level Tile Contextualizer for Histopathology Representation Learning},
  author={Belagali, Varun and Kapse, Saarthak and Marza, Pierre and Das, Srijan and Li, Zilinghan and Boutaj, Sofi{\`e}ne and Pati, Pushpak and Yellapragada, Srikar and Nandi, Tarak Nath and Madduri, Ravi K and others},
  journal={arXiv preprint arXiv:2512.21331},
  year={2025}
}

@inproceedings{ilse2018attention,
  title={Attention-based deep multiple instance learning},
  author={Ilse, Maximilian and Tomczak, Jakub and Welling, Max},
  booktitle={International conference on machine learning},
  pages={2127--2136},
  year={2018},
  organization={PMLR}
}

@misc{chtn,
  title={Cooperative Human Tissue Network ({CHTN})},
  author={{National Cancer Institute}},
  howpublished={\url{https://www.chtn.org}},
  year={2024}
}

@article{campanella2019clinical,
  title={Clinical-grade computational pathology using weakly supervised deep learning on whole slide images},
  author={Campanella, Gabriele and Hanna, Matthew G and Geneslaw, Luke and Miraflor, Allen and Werneck Krauss Silva, Vitor and Busam, Klaus J and Brogi, Edi and Reuter, Victor E and Klimstra, David S and Fuchs, Thomas J},
  journal={Nature medicine},
  volume={25},
  number={8},
  pages={1301--1309},
  year={2019},
  publisher={Nature Publishing Group US New York}
}

@article{runevic2025combining,
  title={Combining Foundation Models in Computational Pathology: Unlocking Multi-Representational Insights},
  author={Runevic, Joel},
  year={2025}
}

@article{kingma2014adam,
  title={Adam: A method for stochastic optimization},
  author={Kingma, Diederik P and Ba, Jimmy},
  journal={arXiv preprint arXiv:1412.6980},
  year={2014}
}

@article{chen2026ranger,
  title={RANGER: Sparsely-Gated Mixture-of-Experts with Adaptive Retrieval Re-ranking for Pathology Report Generation},
  author={Chen, Yixin and Su, Ziyu and Khan, Hikmat and Niazi, Muhammad Khalid Khan},
  journal={arXiv preprint arXiv:2603.04348},
  year={2026}
}

@article{chen2026histomet,
  title={HistoMet: A Pan-Cancer Deep Learning Framework for Prognostic Prediction of Metastatic Progression and Site Tropism from Primary Tumor Histopathology},
  author={Chen, Yixin and Su, Ziyu and Meng, Lingbin and Hasanov, Elshad and Chen, Wei and Parwani, Anil and Niazi, M},
  journal={arXiv preprint arXiv:2602.07608},
  year={2026}
}
\end{spacing}
\end{nolinenumbers}

\end{document}